\documentclass[11pt]{article}

\usepackage[final]{acl}

\usepackage{times}
\usepackage{latexsym}

\usepackage[T1]{fontenc}

\usepackage[utf8]{inputenc}

\usepackage{microtype}

\usepackage{inconsolata}

\usepackage{graphicx}

\usepackage{float}
\usepackage{multirow}
\usepackage{amsmath} 
\usepackage{textcomp}
\usepackage{tabularx}
\usepackage[normalem]{ulem}
\usepackage{booktabs}
\usepackage{footnote}
\usepackage{threeparttable}
\usepackage{booktabs}
\usepackage{caption}
\usepackage{amssymb}             
\usepackage{amsfonts}            
\usepackage{mathrsfs} 
\usepackage{CJKutf8}
\usepackage[table]{xcolor}
\usepackage[noend]{algpseudocode}
\usepackage{booktabs}
\usepackage{indentfirst}
\usepackage{threeparttable}
\usepackage{algorithmicx}
\usepackage{algorithm}
\usepackage{array}
\usepackage{stfloats}
\usepackage{url}
\usepackage{verbatim}
\usepackage{float}
\usepackage{booktabs} 
\usepackage{subfigure}
\usepackage{colortbl}
\usepackage{cleveref}
\usepackage{makecell}
\usepackage{rotating}
\usepackage{arydshln}

\title{Avoiding Over-smoothing in Social Media Rumor Detection with Pre-trained Propagation Tree Transformer}

\author{
 \textbf{Chaoqun Cui},
 \textbf{Caiyan Jia\thanks{Corresponding author.}}
\\
\\
Beijing Key Laboratory of Traffic Data Mining and Embodied Intelligence,\\Beijing Jiaotong University, Beijing 100044, China
\\
 \small{
   \textbf{Correspondence:} \href{ccqun19990728@gmail.com}{ccqun19990728@gmail.com}, \href{cyjia@bjtu.edu.cn}{cyjia@bjtu.edu.cn}
 }
}

\begin{document}
\maketitle
\begin{abstract}

Deep learning techniques for rumor detection typically utilize Graph Neural Networks (GNNs) to analyze post relations. These methods, however, falter due to over-smoothing issues when processing rumor propagation structures, leading to declining performance. Our investigation into this issue reveals that over-smoothing is intrinsically tied to the structural characteristics of rumor propagation trees, in which the majority of nodes are 1-level nodes. Furthermore, GNNs struggle to capture long-range dependencies within these trees. To circumvent these challenges, we propose a Pre-Trained Propagation Tree Transformer (P2T3) method based on pure Transformer architecture. It extracts all conversation chains from a tree structure following the propagation direction of replies, utilizes token-wise embedding to infuse connection information and introduces necessary inductive bias, and pre-trains on large-scale unlabeled datasets. Experiments indicate that P2T3 surpasses previous state-of-the-art methods in multiple benchmark datasets and performs well under few-shot conditions. P2T3 not only avoids the over-smoothing issue inherent in GNNs but also potentially offers a large model or unified multi-modal scheme for future social media research.

\end{abstract}

\section{Introduction}

Currently, deep learning-based methods for rumor detection can be broadly classified into four categories: time-series based methods \cite{weibo,yucnn,liuandwu}, propagation structure learning methods \cite{rvnn,bigcn,ebgcn}, multi-source integration methods \cite{ms1,ms2}, and multi-modal fusion methods \cite{eann,otherrumor1}. Among these, propagation structure learning methods, which leverage the wisdom of crowds to model the relations between posts, have shown promising results in debunking rumors \cite{plan,bigcn,gacl}. 

Numerous studies highlight the value of propagation structures in revealing inter-post relations for rumor detection \cite{gacl,ragcl}. Propagation structure learning methods represent the propagation process of each claim as a tree, which can be a top-down or bottom-up directed or undirected graph with the source post as the root node, comments as other nodes, and reply relations as edges. Graph Neural Networks (GNNs) are then used to learn the representations of rumor propagation trees (RPTs). Two examples of rumor propagation trees are shown in Fig.~\ref{fig:pt}. 


However, our investigation reveals that certain structural characteristics of RPTs pose serious over-smoothing risks for GNNs. Indeed, in our experiments, GNNs showed over-smoothing signs when handling RPTs. This manifests as a decline in the performance of the GNNs when faced with the undirected graph of RPT, as opposed to adopting top-down or bottom-up directed graph. This is particularly contradictory since earlier research \cite{bigcn} has already proven the effectiveness of bidirectional information in RPTs. Moreover, some GNNs, or those dedicated rumor detection models, display a phenomenon where the model's accuracy decreases as the model depth increases. These experimental phenomena demonstrate that GNNs tend to encounter over-smoothing problems when dealing with RPTs. It especially hinders the expansion of model scale during pre-training on large-scale unlabeled social media data.

We identified the causes of over-smoothing by investigating RPT structures. We found that in the propagation process of social posts, the vast majority of comments are direct replies to source post. This results in the majority of nodes in a RPT being 1-level nodes directly connected to the root, while the proportion of nodes at deeper levels is quite small. Because the scope of a node's neighborhood view in GNNs is tied to the model depth \cite{gcn,gin}, all 1-level nodes in a tree-like structure like a RPT are in each other's 2-hop neighborhoods. If neighborhood aggregation is conducted indiscriminately, it leads to over-smoothing. In other words, the structure of graphs like RPTs inherently predisposes GNNs to over-smoothing.
In addition, GNNs struggle to capture long-range dependencies \cite{long} in RPTs, making it challenging to leverage information from the few deep nodes in the tree. Specifically, GNNs find it hard to learn from complete conversation chains in RPTs (conversation threads from 1-level node to leaf node), and such chains are better suited for Transformer \cite{transformer} architecture designed for sequential structure. As a result, we chose Transformer as underlying architecture for our model to harness its self-attention mechanism, avoiding potential over-smoothing issues and facilitating modeling of user interactions throughout complete conversation chains.

In this study, we propose \textbf{P}re-\textbf{T}rained \textbf{P}ropaga- tion \textbf{T}ree \textbf{T}ransformer (P2T3). P2T3 extracts conversation chains from deep structure of RPTs following the node propagation direction to form sequential structures. It uses token-wise embedding to infuse connection information into the sequence tokens, introduces necessary inductive bias, and pre-trains on large-scale unlabeled datasets to enhance the performance. Experiments shows that P2T3 outperforms previous state-of-the-art (SOTA) methods on multiple benchmark datasets. 

In summary, this study contributes as follows:
\begin{itemize}
\item We ran extensive experiments to reveal over-smoothing phenomenon and its intrinsic causes on RPTs.
\item We released two large unlabeled topic datasets, which may promote semi-supervised rumor detection research.
\item We proposed the P2T3 method, where the special token-wise embedding enables Transformer to handle RPTs. 
\item Experiments show that P2T3 outperforms current SOTA methods and performs well in few-shot scenarios.
\end{itemize}

\section{Empirical Investigation}

\subsection{Over-smoothing Phenomenon}

The over-smoothing phenomena of GNNs when handling RPTs mainly manifest in two aspects: (1) Useing undirected graphs of RPTs to learn bidirectional propagation information actually impedes the improvement of model performance; (2) The expansion of the model scale, by increasing in the number of model layers, leads to a performance decrease. We will elaborate on these two aspects.

\subsubsection{Impact of Infomation Flow Direction}

\begin{figure*}[h]
  \centering
  \subfigure[Weibo]{\includegraphics[width=0.245\textwidth]{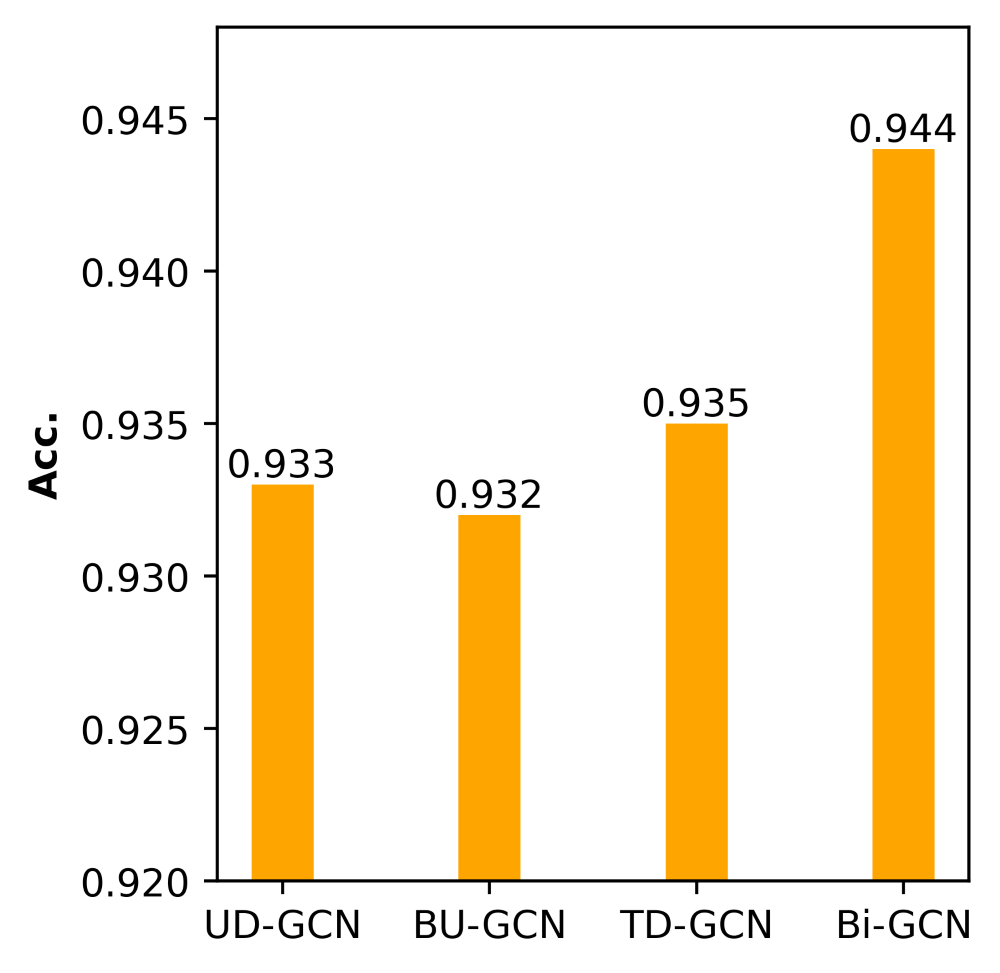}}
  \subfigure[DRWeibo]{\includegraphics[width=0.245\textwidth]{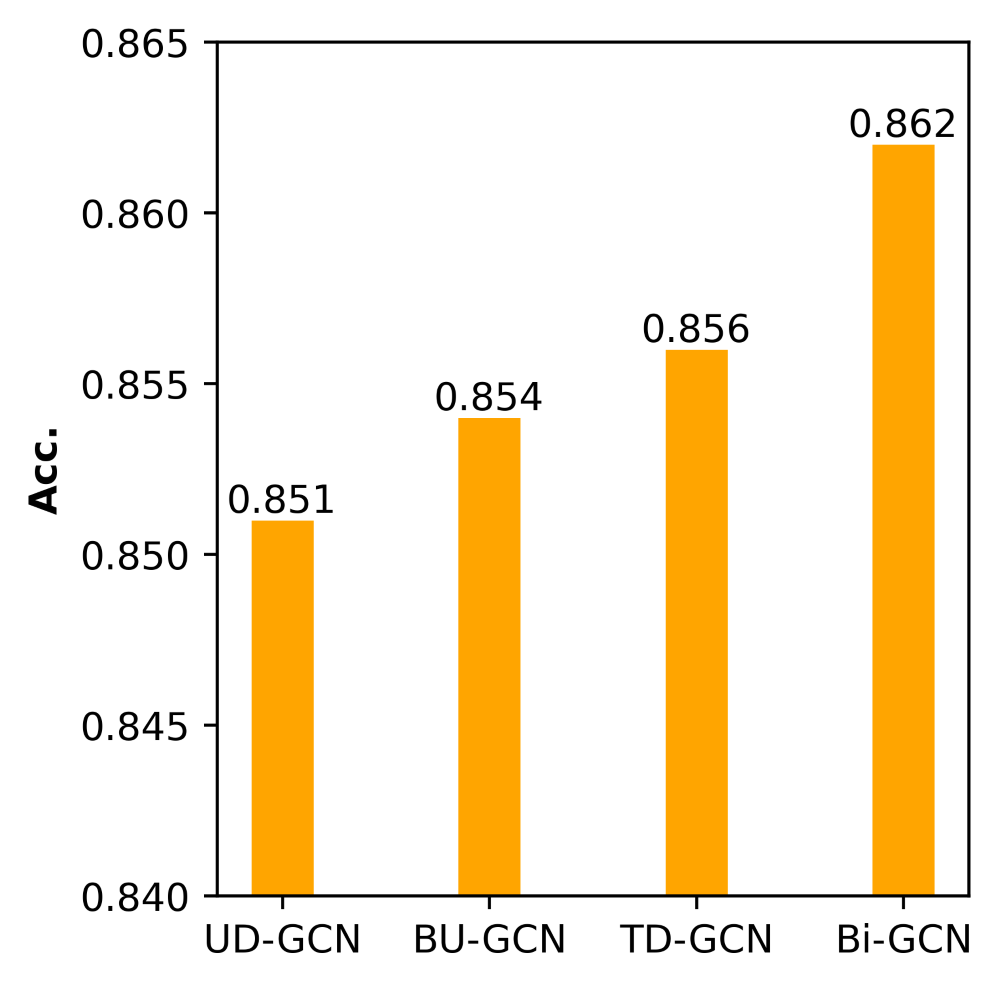}}
  \subfigure[Twitter15]{\includegraphics[width=0.245\textwidth]{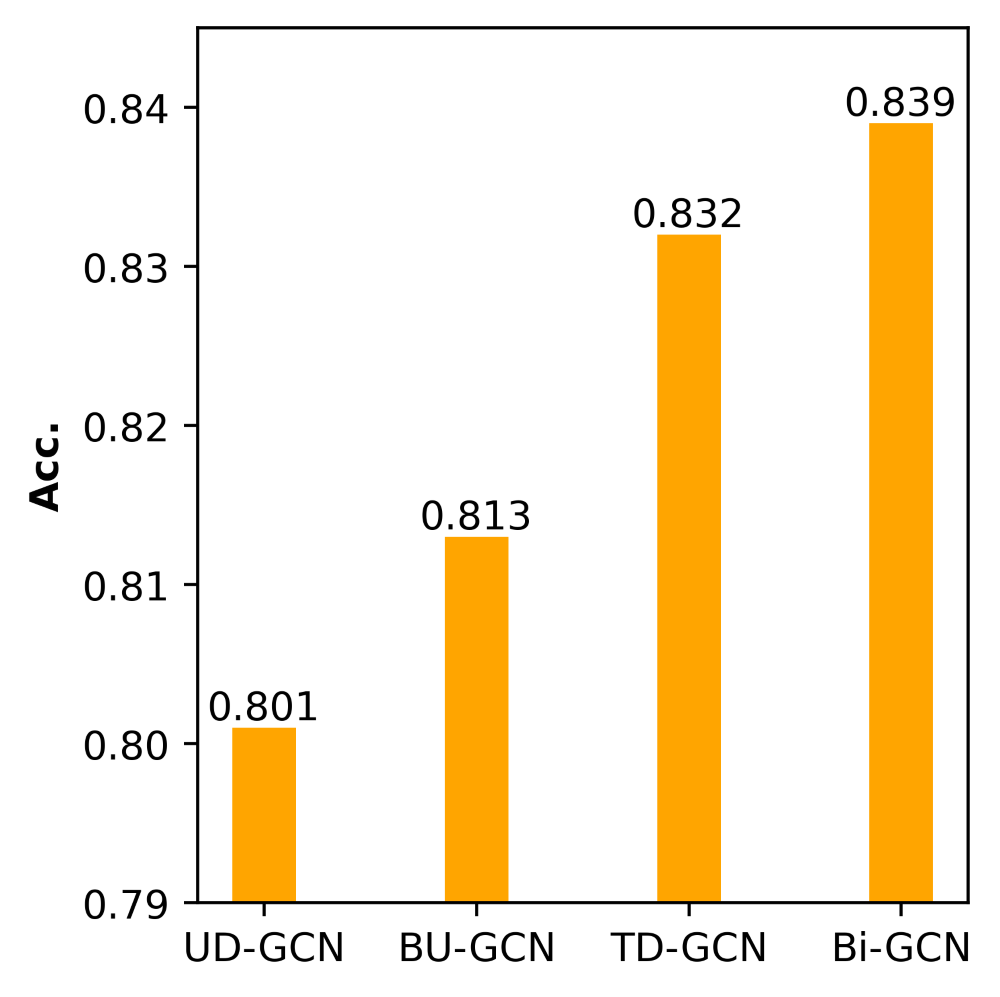}}
  \subfigure[Twitter16]{\includegraphics[width=0.245\textwidth]{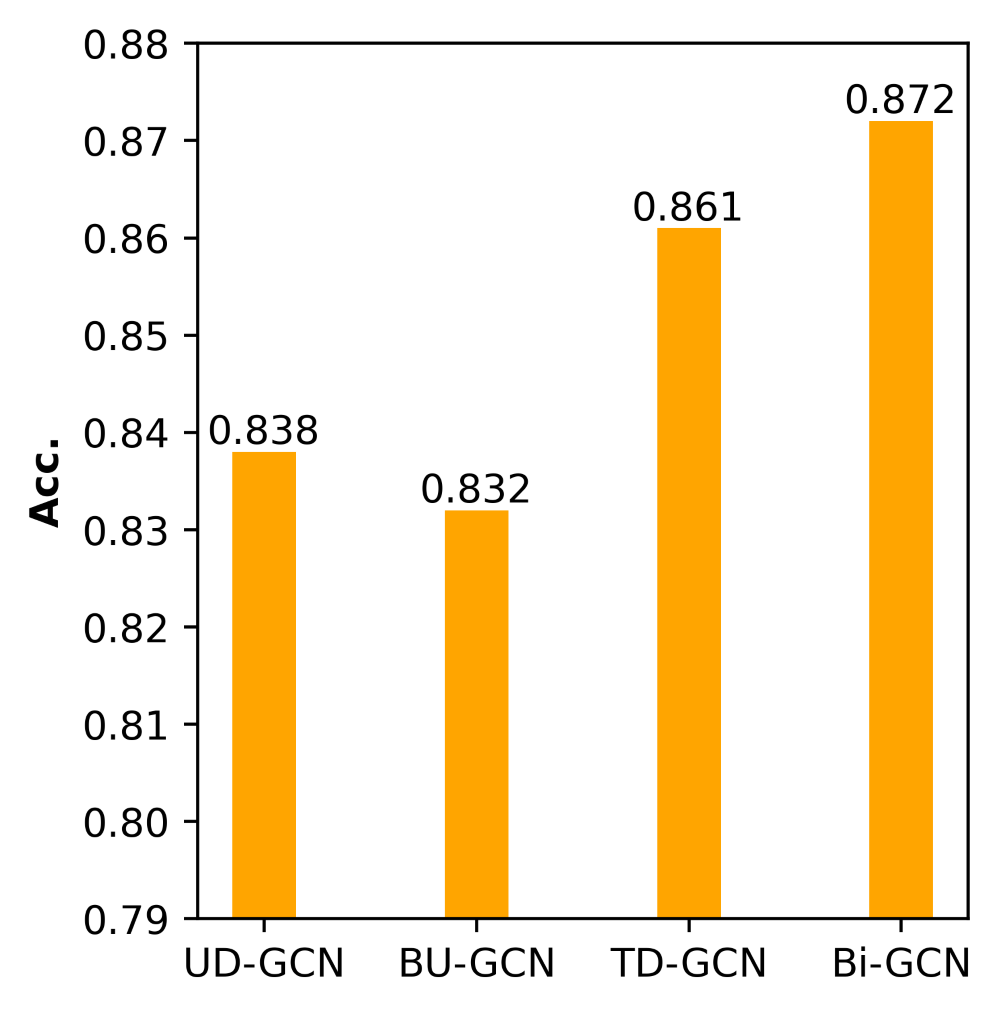}}
  \caption{The impact of information flow in different directions.}
  \label{fig:directional}
\end{figure*}

We conducted experiments on four rumor detection datasets including Weibo \cite{weibo}, DRWeibo \cite{ragcl}, Twitter15, and Twitter16 \cite{twitter1516}, following Bian et al. (2020) \cite{bigcn}. We employed GCNs in four settings: undirected graphs, top-down directed graphs, bottom-up directed graphs, and a combination of top-down and bottom-up graphs, labeled as UD-GCN, TD-GCN, BU-GCN, and Bi-GCN respectively. UD-GCN uses one GCN encoder, while Bi-GCN uses two GCN encoders on top-down and bottom-up graphs. Results is shown in Fig.~\ref{fig:directional}.

The results reveal two phenomena: (1) Bi-GCN shows a significant improvement compared to TD-GCN and BU-GCN, which only use top-down or bottom-up information. (2) UD-GCN does not show a substantial improvement over TD-GCN and BU-GCN, and performance even decreases on certain datasets (DRWeibo and Twitter15). The improved performance of Bi-GCN compared to TD-GCN and BU-GCN suggests that both top-down and bottom-up information in RPTs are useful. However, the effect of using UD-GCN to learn this bidirectional information is not ideal and may even result in performance degradation. These phenomena reflect a contradictory fact, namely, while bidirectional information is beneficial, learning from undirected graphs can lead to adverse effects. The ablation studies by BiGCN \cite{bigcn} and RAGCL \cite{ragcl} also demonstrate similar phenomena.

\subsubsection{Deep Model Performance Degradation}

\begin{figure*}[h]
  \centering
  \subfigure[Weibo]{\includegraphics[width=0.245\textwidth]{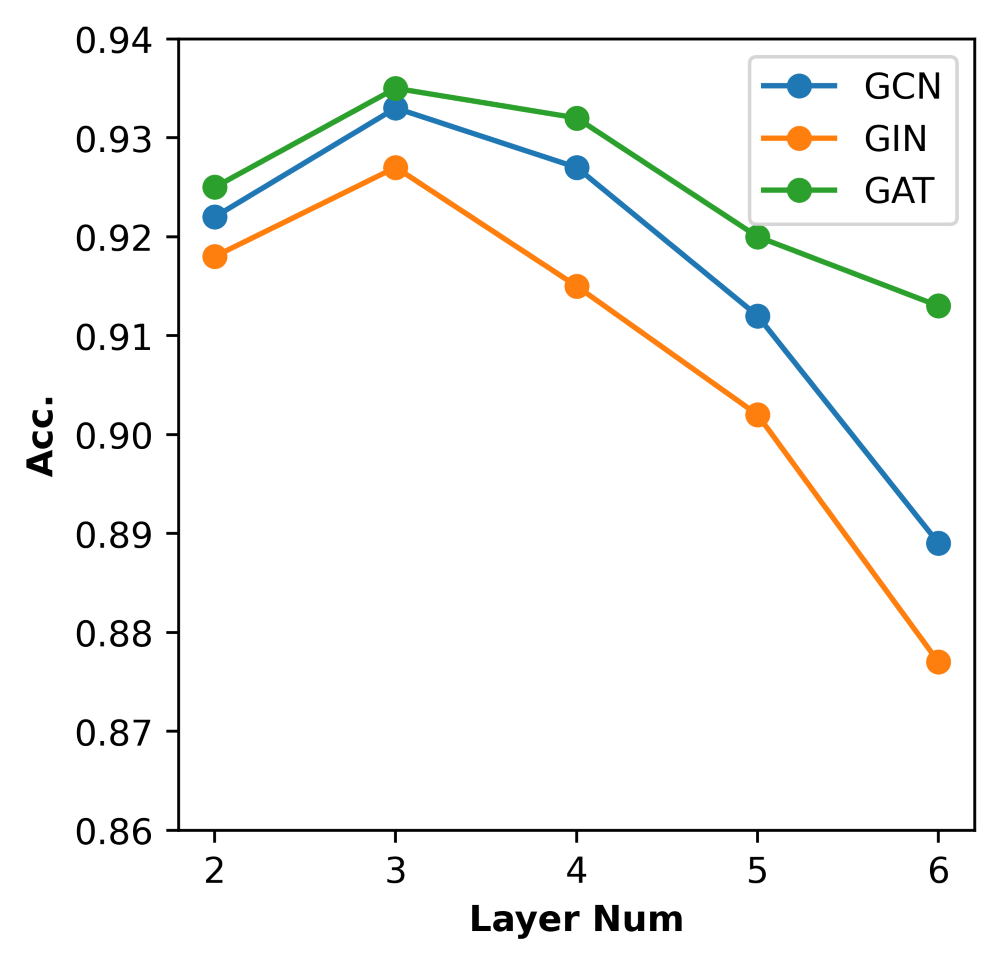}}
  \subfigure[DRWeibo]{\includegraphics[width=0.245\textwidth]{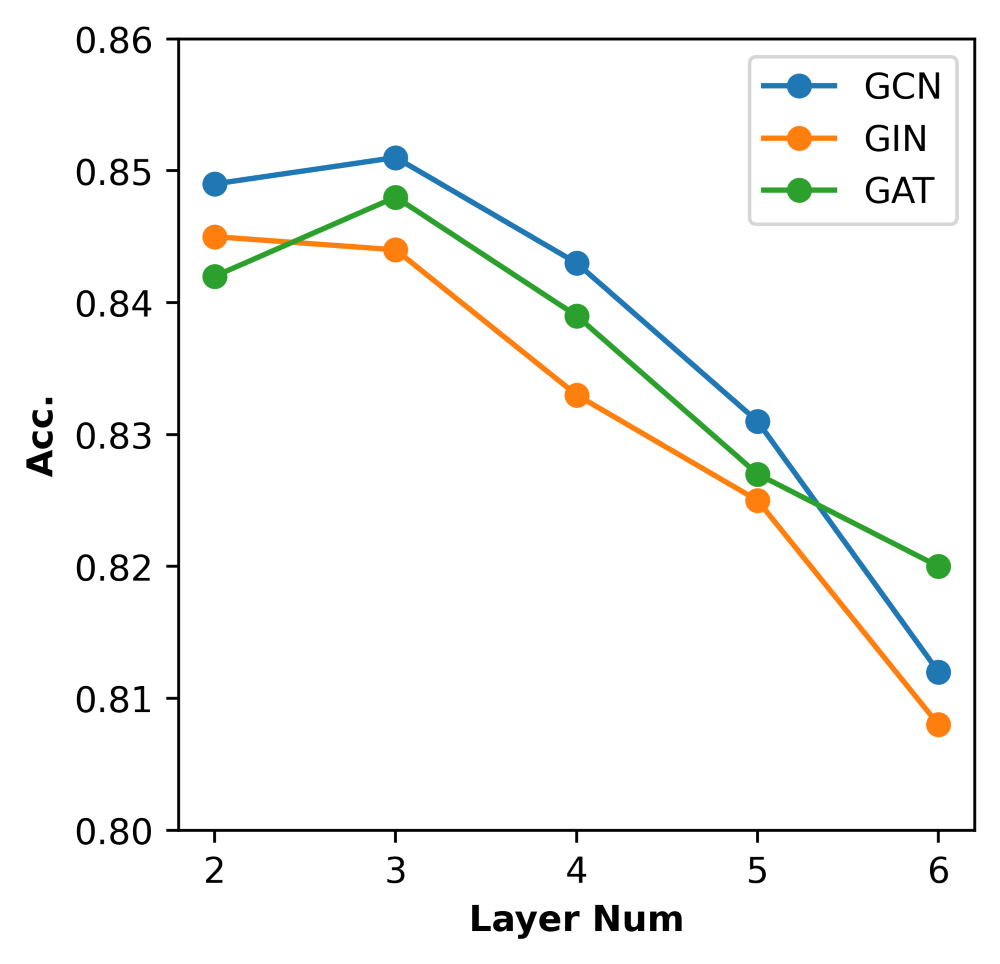}}
  \subfigure[Twitter15]{\includegraphics[width=0.245\textwidth]{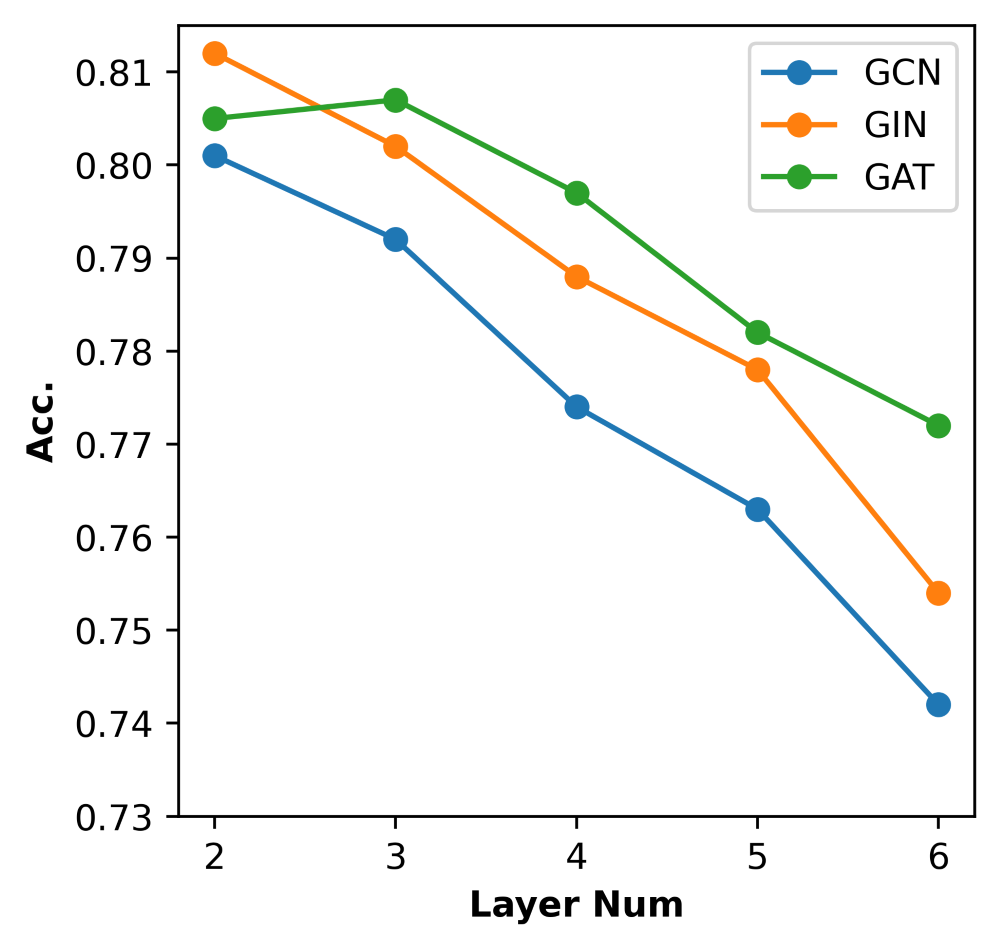}}
  \subfigure[Twitter16]{\includegraphics[width=0.245\textwidth]{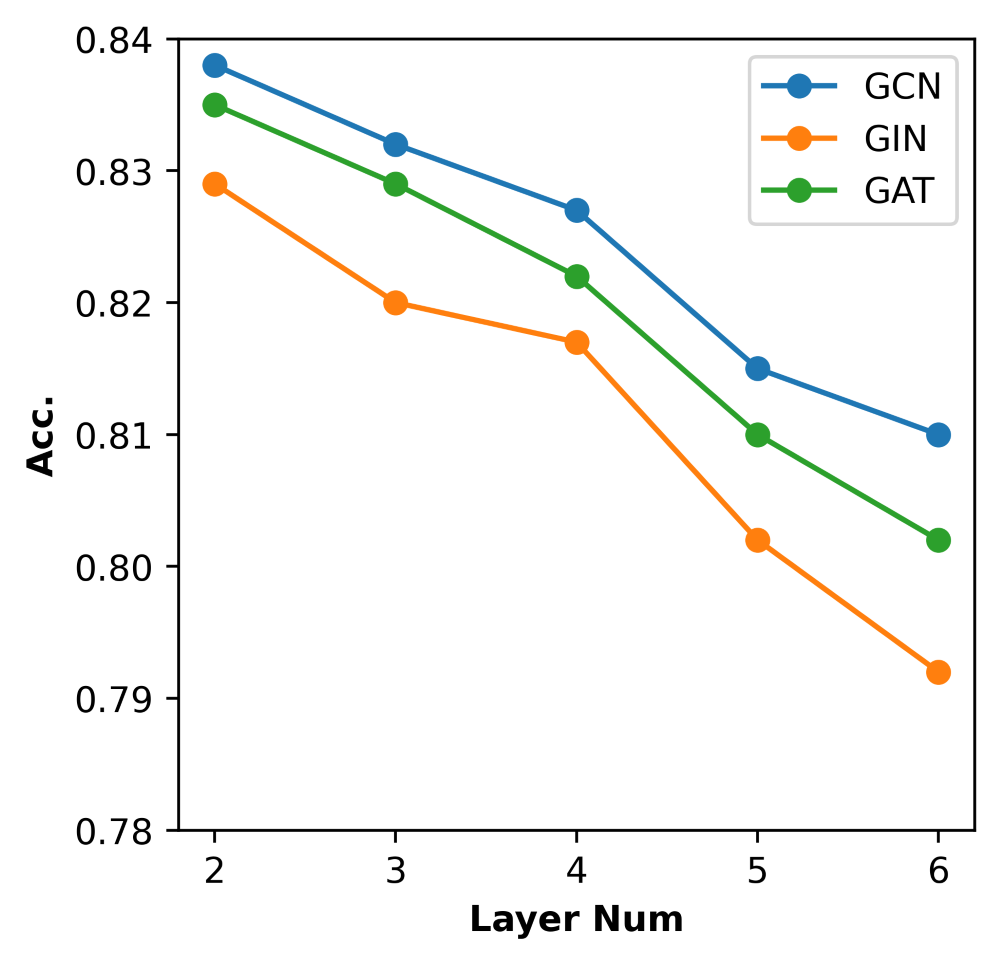}}
  \caption{The impact of model layer numbers.}
  \label{fig:layernum}
\end{figure*}

We explored the effect of varying GNN layer numbers on performance using undirected RPT graphs (see Fig.~\ref{fig:layernum}). It suggests that the optimal GNN layer count for Weibo and DRWeibo is three, while it's two for Twitter15 and Twitter16. Performance declines to different degrees as layer number increases. In other words, the model scale expansion actually hinders model performance improvement, which is a typical over-smoothing phenomenon \cite{oversm1,oversm2}. Among the three GNNs, GAT seems more robust to model depth, possibly due to its attention mechanism that adaptively aggregates neighborhoods, thereby mitigating over-smoothing.

\subsection{Theoretical Cause Analysis}

Above phenomena shows that over-smoothing occurs when GNNs deal with RPTs, which makes it difficult for GNNs to effectively utilize bidirectional propagation information. Furthermore, it especially hinders the expansion of model scale during pre-training on large-scale social media data. We believe that the occurrence of this over-smoothing phenomenon on RPTs is not only due to the limitations of GNNs, but also the inherent structural characteristics of these trees naturally lead to over-smoothing. Specifically, we have collected statistics on the depth distribution of posts across multiple datasets showed in Table~\ref{tab:structure}.

\begin{table*}[h]
\centering
\resizebox{\textwidth}{!}{
\begin{tabular}{ccccc|cc}
\Xhline{1.0pt}
\rowcolor{gray!20}
\textbf{Statistic} & \textbf{Weibo} & \textbf{DRWeibo} & \textbf{Twitter15} & \textbf{Twitter16} & \textbf{UWeibo} & \textbf{UTwitter}\\
\hline
\textbf{\# claims} & 4664 & 6037 & 1490 & 818 & 209549 & 204922\\
\textbf{\# non-rumors} & 2351 & 3185 & 374 & 205 & - & -\\
\textbf{\# false rumors} & 2313 & 2852 & 370 & 205 & - & -\\
\textbf{\# true rumors} & - & - & 372 & 207 & - & -\\
\textbf{\# unverified rumors} & - & - & 374 & 201 & - & -\\
\hdashline
\textbf{\# avg reply} & 803.5 & 61.8 & 50.2 & 49.1 & 50.5 & 82.5\\
\textbf{\# avg 1-level reply} & 522.9(65\%) & 48.1(78\%) & 35.5(71\%) & 31.6(64\%) & 36.4(72\%) & 48.5(59\%)\\
\textbf{\# avg 2-level reply} & 169.3(21\%) & 11.0(17\%) & 5.9(12\%) & 6.0(12\%) & 10.2(20\%) & 21.5(26\%)\\
\textbf{\# avg deeper reply} & 111.2(14\%) & 2.7(5\%) & 8.7(17\%) & 11.5(24\%) & 4.0(8\%) & 12.5(15\%)\\
\Xhline{1.0pt}
\end{tabular}
}
\caption{Statistics of the datasets.}
\label{tab:structure}
\end{table*}

Statistics reveal that RPTs are highly imbalanced, with dense connections at central node (root) and sparse connections at deeper nodes. This leads to a situation where even a very shallow GNN can enable nodes to reach almost all other nodes when learning undirected graphs of RPTs. For example, a 2-layer GNN lets 1-level nodes gather information from all other 1-level nodes, and a 3-layer GNN allows 2-level nodes to reach all 1-level nodes. For datasets with small average depth (e.g., DRWeibo, Twitter15, Twitter16, etc.), when using a shallow GNN, nodes can almost aggregate information from all nodes in the graphs during forward propagation process. This is an important reason for over-smoothing in rumor detection models. 

Because the neighborhood view scope of a node is tied to GNN depth, and information propagates along edge direction \cite{gcn,graphsage}, indiscriminately aggregating node neighborhoods when GNNs face undirected graphs of RPTs can lead to over-smoothing. However, this problem of excessively broad node view doesn't occur when using directed graphs, which explains why sometimes better performance can be achieved with top-down or bottom-up directed graphs.

From the spectral graph theory view, taking GCN as an example, a GCN essentially performs a low-pass filtering on eigenvectors of graph Laplacian matrix \cite{gcn}. Low-frequency components are preserved, while high-frequency components are filtered out. For RPTs, since most nodes are leaf node and directly connected to roots, the frequency is predominantly concentrated on low-frequency range, with scant information in high-frequency range. In GCN, high-frequency information (such as features of the deeper nodes which are believed to contain discriminative features for debunking rumors \cite{ragcl}) may be excessively filtered out, thereby failing to utilize full conversation chain information and leading to over-smoothing when GCNs aggregate neighborhood, due to the imbalance of the node distribution within RPTs. 

\section{Method}

\subsection{Conversation Chains in RPTs}

In RPT, each conversation thread from 1-level node to each leaf node forms a separate conversation chain, which can be considered as a sequential chain-like structure (see Fig.~\ref{fig:input}). Such a conversation chain represents a complete conversation group during the propagation process of a claim, with the nodes in the conversation arranged in chronological order. There are explicit semantic relations and stance expressions between the nodes. In general, nodes in longer conversation chains tend to express stronger sentiments. This is because the individuals participating in these conversations are often engaged in heated arguments or debates regarding a certain topic. If a model can learn from these conversation chains, it would be helpful for rumor debunking. However, when GNNs handle RPTs, they focus more on all direct replies to a node and cannot learn information from a complete conversation chain. This is because GNNs follow a neighborhood aggregation framework, which emphasizes breadth of information compared to Transformer architecture and cannot capture long-range dependencies from deep conversation chains. This is another important reason why we chose Transformer architecture.
In Tables~\ref{tab:cc}, we present several conversation chains with their source posts. It's evident that these chains, often chronologically arranged, exhibit clear semantic relations and occasionally intense emotional expressions. These are notable features for identifying rumors.

\subsection{Input Representation}

\begin{figure*}[t]
  \centering
  \includegraphics[width=0.98\textwidth]{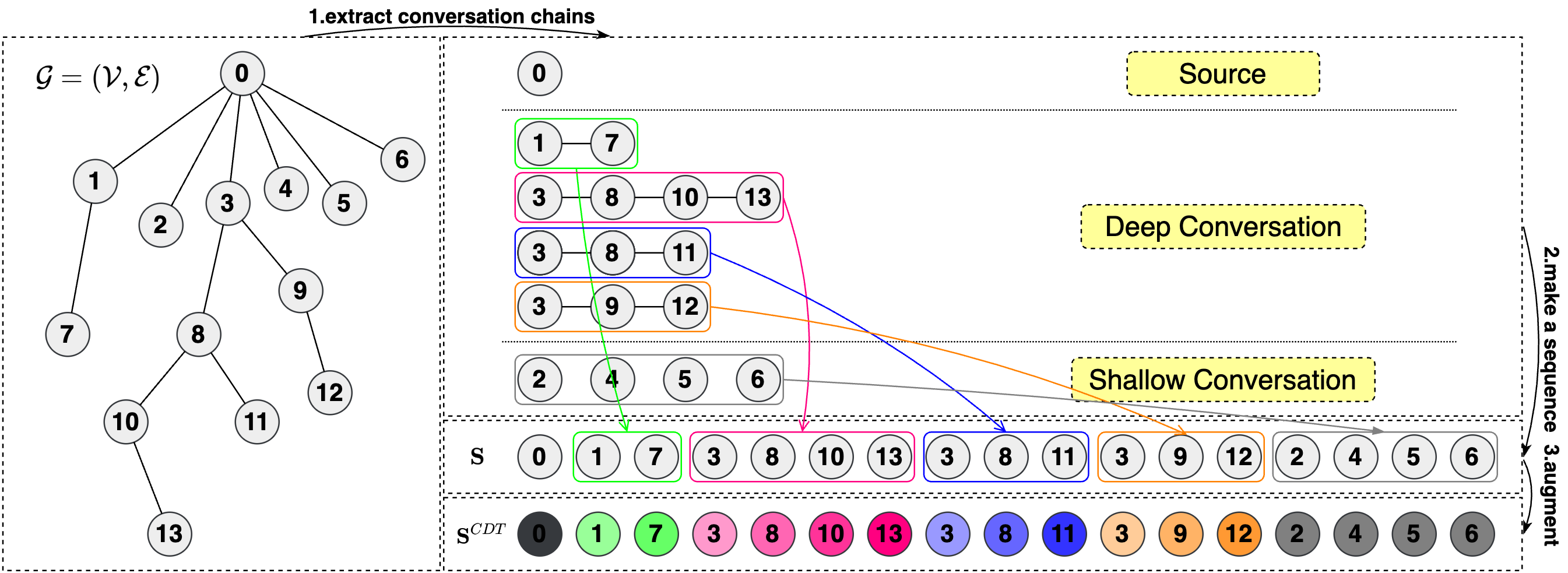}
  \caption{The input of P2T3. Different colors represent chain identifiers. Color shades represent depth embeddings.}
  \label{fig:input}
\end{figure*}

P2T3 adopts Transformer over GNNs primarily for the following reasons. (1) The Graph Transformer, by allowing nodes to attend to all other nodes (global attention), alleviates fundamental limitations of sparse message passing mechanism, such as over-smoothing and limited expressiveness. (2) Nodes in RPTs spread directionally according to a top-down distribution, demonstrating a canonical node ordering. 
Transformers allows to extract sequential structures like conversation chains, easily, thus enables to learn rumor patterns from these sequential structures by capturing long-range dependencies.
\citet{tokengt} showed that Transformer can recognize graph connectivity through suitable token-wise embeddings. We were inspired to design specific embeddings for RPTs. We first extract all conversation chains from a tree, classify the conversations into three types: source, deep conversation, and shallow conversation, then combine them into a sequence. Their features are augmented with token-wise embeddings. Through specific embedding designs, unimportant connections are downplayed, mitigating RPT's over-smoothing issue. This process is shown in Fig.~\ref{fig:input}.

For a RPT $\mathcal{G}=(\mathcal{V},\mathcal{E})$, where $\mathcal{V}$ and $\mathcal{E}$ are sets of nodes and edges, $\mathbf{X}\in \mathbb{R}^{n\times d}$ represents node feature vectors, with dimension $d$. We first extract conversation chains from $\mathcal{G}$ and compose them into a sequence $\mathbf{S}\in \mathbb{R}^{m\times d}$, where $m\geq n$. For each $v\in \mathcal{V}$, it may appear multiple times in $\mathbf{S}$. The tokens in $\mathbf{S}$ is denoted as $\mathcal{T}=\left \{t_{1},\cdots ,t_{m}\right \}$. We use three token-wise embeddings to augment $\mathbf{S}$: chain identifier, depth embedding, and type embedding. These embeddings are used to assist the model in identifying the ownership of chains in the sequence, determining the depth of nodes in the tree, and capturing the conversation types, respectively.

\noindent \textbf{Chain Identifier.} For a RPT $\mathcal{G}=(\mathcal{V},\mathcal{E})$, we provide $l$ orthonormal vectors $\mathbf{C}\in \mathbb{R}^{l\times l}$ ($l\geq m$). Tokens within $\mathbf{S}$ belonging to the same conversation are assigned identical vectors, while tokens from different conversations receive distinct vectors. This ensures that for any two tokens $u$ and $v$ in $\mathcal{T}$, $\mathbf{C}_{u}\mathbf{C}_{v}^{T}=1$ only if $u$ and $v$ belong to the same conversation; otherwise, it is 0. Specifically, for each token $t\in \mathcal{T}$ in $\mathbf{S}$, we augment $\mathbf{S}_{t}$ by concatenating it with $\mathbf{C}_{t}$, resulting in $[\mathbf{S}_{t},\mathbf{C}_{t}]$. Then, we apply a parameter matrix $w\in \mathbb{R}^{(d+l)\times d}$ to map the augmented sequence back to the $d$-dimensional space, obtaining the sequence $\mathbf{S}^{C}\in \mathbb{R}^{m\times d}$. Notably, as the chain identifier matrix $\mathbf{C}$ is only required to be orthonormal, we use the matrix $\mathbf{Q}\in \mathbb{R}^{l\times l}$ obtained by performing QR decomposition on random Gaussian matrix $\mathbf{G}\in \mathbb{R}^{l\times l}$ as the matrix $\mathbf{C}$ in practice.

\noindent \textbf{Depth Embedding.} The depth relation of nodes in a conversation chain is akin to the positional relation in a sequence. Consequently, any positional encoding of the Transformer model \cite{transformer,rope} is compatible. In P2T3, we simply adopt an approach resembling sinusoidal positional embeddings to represent the depth of tokens in $\mathcal{T}$ within the original RPT: 
\begin{equation}
\begin{split}
&\mathbf{D}_{dph,2i}=sin(dph/10000^{2i/d}),\\
&\mathbf{D}_{dph,2i+1}=cos(dph/10000^{2i/d}),
\end{split}
\end{equation}
where $dph$ is the depth level and $i$ is the dimension. This facilitates the model in recognizing reply relations within conversation chains. Then, we utilize these depth embeddings to augment the tokens in $\mathbf{S}^{C}$. Specifically, for each token $t\in \mathcal{T}$ in $\mathbf{S}^{C}$, we add $\mathbf{D}_{dph(t)}$ to $\mathbf{S}_{t}^{C}$, resulting in $\mathbf{S}_{t}^{C}+\mathbf{D}_{dph(t)}$. Here, $dph(t)$ represents the original depth of token $t$ in the tree. The augmented $\mathbf{S}^{C}$ is denoted as $\mathbf{S}^{CD}$.

\noindent \textbf{Type Embedding.} Type embedding is used to identify the conversation type that each token belongs to. The type embedding is a $d$-dimensional vector of all 0s, all 1s, or all 2s, representing respectively that the token belongs to the source, deep conversation, or shallow conversation types. Specifically, for each token $t\in \mathcal{T}$ in $S^{CD}$, we add $\mathbf{S}_{t}^{CD}$ and type embedding $\mathbf{T}_{tp(t)}$ to get $\mathbf{S}_{t}^{CD}+\mathbf{T}_{tp(t)}$. Here, $tp(t)$ represents the conversation type that $t$ corresponds to. The $\mathbf{S}^{CD}$ augmented by type embedding is denoted as $\mathbf{S}^{CDT}$, and $\mathbf{S}^{CDT}$ is the input of standard Transformer encoder.

\subsection{Training Strategy}
 
P2T3 pre-trains on massive unlabeled dataset, then fine-tunes on labeled dataset. We built two large unlabeled datasets, UWeibo and UTwitter, for rumor detection on Weibo and Twitter. Each contains over 200,000 claims from social platforms. Each claim includes source post and replies, as well as propagation structure. These datasets are available at \url{https://anonymous.4open.science/r/UWeibo-D405} and \url{https://anonymous.4open.science/r/UTwitter-C882}. 

\noindent \textbf{Pre-training.} Rumor detection tasks focus on the interaction between source posts and their replies. Therefore, in the pre-training process, we maximize the Mutual Information (MI) \cite{dgi,infograph} between a source post in $\mathcal{T}$ and the first token of all conversations (i.e., all 1-level nodes in the RPT):
\begin{equation}
\small
\mathcal{L}_{unsup}(\mathbb{S}^{U})=-\frac{1}{|\mathbb{S}^{U}|}\sum _{\mathbf{S}\in \mathbb{S}^{U}}\sum _{n\in \mathcal{N}}I(h_{n}(\mathbf{S});h_{root}(\mathbf{S})),
\end{equation}
where $\mathbb{S}^{U}$ is input sequences set from unlabeled samples, $\mathcal{N}$ is set of first tokens of all conversations in $\mathbf{S}$, $h_{root}(\mathbf{S})$ is source post representation, $I(\cdot ;\cdot )$ denotes MI contrastive loss between two representations. There are various methods available for computing MI, such as Donsker-Varadhan representation, Jensen-Shannon MI estimator, InfoNCE, etc \cite{dim}. We employ such loss to enhance the consistency between source post and its replies, promoting alignment between global and local representations, enabling the model to fully learn user interactions and emotional expressions.

\noindent \textbf{Fine-tuning.} After pre-training, we can fine-tune the model on a supervised rumor detection dataset. Specifically, for each sequence $\mathbf{S}$ in the labeled dataset $\mathbb{S}^{L}$, we pass the source post representation $h_{root}(\mathbf{S})$ through a fully connected classifier, and then compute the cross-entropy loss:
\begin{equation}
\small
\mathcal{L}_{sup}(\mathbb{S}^{L})=-\frac{1}{|\mathbb{S}^{L}|}\sum _{\mathbf{S}\in \mathbb{S}^{L}}CE(f(h_{root}(\mathbf{S})),y),
\end{equation}
where $y$ is ground truth label, $f(\cdot )$ is classifier, $CE(\cdot ,\cdot )$ is cross-entropy loss. Optionally, the unsupervised loss $\mathcal{L}_{unsup}(\mathbb{S}^{L})$ can be weighted and added to $\mathcal{L}_{sup}(\mathbb{S}^{L})$.

\section{Experiments}




\subsection{Experimental Settings}

We introduce some specific settings in this subsection, including baselines we compared, procedure of data preprocessing and hyperparameter configuration when training the model. The source code of P2T3 is available at \url{https://anonymous.4open.science/r/P2T3-E83D}.

\subsubsection{Baselines}

We ran experiments on 4 real-world benchmark datasets (see Table~\ref{tab:structure}). Weibo and DRWeibo are used with UWeibo. Twitter15 and Twitter16 are with UTwitter. 

We compare with the following baseline methods: \textbf{PLAN} \cite{plan}, \textbf{HD-TRANS} \cite{treetrans}, \textbf{BiGCN} \cite{bigcn}, \textbf{ClaHi-GAT} \cite{clahi}, \textbf{GACL} \cite{gacl}, \textbf{DDGCN} \cite{ddgcn}, and \textbf{RAGCL} \cite{ragcl}.

\subsection{Results and Discussion}

The main results are shown in Tables~\ref{tab:weibo} and~\ref{tab:twitter}. P2T3 outperforms baselines on all datasets.
PLAN and HD-TRANS, two Transformer-based methods, exhibit inferior performance compared to GNN-based approaches, possibly due to the inappropriate ways in which they fuse propagation structure information. In contrast, P2T3 leverages the same Transformer architecture but achieves superior results, indicating the efficacy of our token-wise embedding and the importance of learning from conversation chains.
BiGCN is a typical model built on the deep structure of RPT, which presupposes that the information flow in RPTs presents as a top-down propagation and a bottom-up dispersion process. However, our investigation indicate that RPT actually manifests as a shallow imbalanced structure. This imbalanced distribution of information in the depth direction is also an important characteristic, which is overlooked by existing techniques. Perhaps as a result of this observation, P2T3 achieves a performance boost over BiGCN.
ClaHi-GAT uses a gating module to filter neighborhood information while integrating sibling connections into the undirected graph. Additionally, it utilizes a shallow GAT architecture that is less affected by over-smoothing (see Fig.~\ref{fig:layernum}). As a result, it is not significantly impacted by over-smoothing. However, its overall performance remains noticeably inferior to P2T3.
Although GACL uses BERT \cite{bert} for initial feature extraction, its improvement is marginal. This may suggest that rumor detection models are insensitive to the way initial features are extracted, and what is more crucial is the high-level model's ability to learn node interactions. 


\begin{table*}[!h]
\centering
\resizebox{0.98\textwidth}{!}{
\begin{tabular}{cccccccccc}
 \Xhline{1.0pt}
 \rowcolor{gray!20}
 ~ & ~ & \multicolumn{4}{c}{\textbf{Weibo}} & \multicolumn{4}{c}{\textbf{DRWeibo}}\\
 \cline{3-10}
 \rowcolor{gray!20}
 \multirow{-2}{*}{\textbf{Method}} & \multirow{-2}{*}{\textbf{Class}} & \textbf{Acc.} & \textbf{Prec.} & \textbf{Rec.} & \textbf{F1} & \textbf{Acc.} & \textbf{Prec.} & \textbf{Rec.} & \textbf{F1}\\
 \hline
 \multirow{2}{*}{PLAN} & R & \multirow{2}{*}{0.915\small ±0.007} & 0.908 & 0.923 & 0.915 & \multirow{2}{*}{0.788\small ±0.005} & 0.786 & 0.760 & 0.771 \\
 & N & ~ & 0.923 & 0.907 & 0.914 & ~ & 0.793 & 0.813 & 0.802 \\
 \hline
 \multirow{2}{*}{HD-TRANS} & R & \multirow{2}{*}{0.921\small ±0.004} & 0.915 & 0.929 & 0.920 & \multirow{2}{*}{0.810\small ±0.006} & 0.814 & 0.834 & 0.823\\
 & N & ~ & 0.928 & 0.913 & 0.921 & ~ & 0.809 & 0.783 & 0.794 \\
 \hline
 \multirow{2}{*}{BiGCN} & R & \multirow{2}{*}{0.942\small ±0.008} & 0.919 & 0.968 & 0.942 & \multirow{2}{*}{0.866\small ±0.010} & 0.869 & 0.849 & 0.858 \\
 & N & ~ & 0.967 & 0.918 & 0.942 & ~ & 0.863 & 0.882 & 0.872 \\
 \hline
\multirow{2}{*}{ClaHi-GAT} & R & \multirow{2}{*}{0.935\small ±0.009} & 0.932 & 0.934 & 0.933 & \multirow{2}{*}{0.864\small ±0.012} & 0.868 & 0.876 & 0.872 \\
 & N & ~ & 0.938 & 0.936 & 0.937 & ~ & 0.859 & 0.850 & 0.854 \\
 \hline
 \multirow{2}{*}{GACL} & R & \multirow{2}{*}{0.938\small ±0.006} & 0.936 & 0.940 & 0.938 & \multirow{2}{*}{0.870\small ±0.009} & 0.865 & 0.856 & 0.860 \\
 & N & ~ & 0.940 & 0.936 & 0.938 & ~ & 0.874 & 0.882 & 0.878 \\
 \hline
 \multirow{2}{*}{DDGCN} & R & \multirow{2}{*}{0.948\small ±0.004} & 0.924 & \textbf{0.979} & 0.951 & \multirow{2}{*}{0.878\small ±0.005} & 0.872 & 0.864 & 0.868 \\
 & N & ~ & \textbf{0.976} & 0.917 & 0.946 & ~ & 0.883 & 0.891 & 0.887 \\
 \hline
 \multirow{2}{*}{RAGCL} & R & \multirow{2}{*}{0.960\small ±0.006} & 0.954 & 0.972 & 0.959 & \multirow{2}{*}{0.896\small ±0.005} & 0.895 & 0.880 & 0.884 \\
 & N & ~ & 0.967 & 0.959 & 0.962 & ~ & 0.897 & 0.910 & 0.907 \\
 \hline
 \multirow{2}{*}{\textbf{P2T3}} & R & \multirow{2}{*}{\textbf{0.973\small ±0.003}} & \textbf{0.966} & 0.975 & \textbf{0.970} & \multirow{2}{*}{\textbf{0.912\small ±0.005}} & \textbf{0.906} & \textbf{0.900} & \textbf{0.903} \\
 & N & ~ & 0.975 & \textbf{0.969} & \textbf{0.972} & ~ & \textbf{0.913} & \textbf{0.927} & \textbf{0.920} \\
 \Xhline{1.0pt}
\end{tabular}
}
\caption{Experimental results on Weibo and DRWeibo dataset.}
\label{tab:weibo}
\end{table*}

\begin{table*}[!h]
\centering
\resizebox{0.98\textwidth}{!}{
\begin{tabular}{ccccccccccc}
 \Xhline{1.0pt}
 \rowcolor{gray!20}
 ~ & \multicolumn{5}{c}{\textbf{Twitter15}} & \multicolumn{5}{c}{\textbf{Twitter16}}\\
 \cline{2-11}
 \rowcolor{gray!20}
 ~ & ~ & \textbf{N} & \textbf{F} & \textbf{T} & \textbf{U} & ~ & \textbf{N} & \textbf{F} & \textbf{T} & \textbf{U}\\
 \cline{3-6}
 \cline{8-11}
 \rowcolor{gray!20}
 \multirow{-3}{*}{\textbf{Method}} & \multirow{-2}{*}{\textbf{Acc.}} & \textbf{F1} & \textbf{F1} & \textbf{F1} & \textbf{F1} & \multirow{-2}{*}{\textbf{Acc.}} & \textbf{F1} & \textbf{F1} & \textbf{F1} & \textbf{F1}\\
 \hline
 PLAN & 0.819\small ±0.004 & 0.839 & 0.854 & 0.817 & 0.759 & 0.843\small ±0.005 & \textbf{0.855} & 0.851 & 0.858 & 0.805 \\
 HD-TRANS & 0.810\small ±0.008 & 0.834 & 0.841 & 0.750 & 0.808 & 0.828\small ±0.004 & 0.836 & 0.840 & 0.843 & 0.790 \\
 BiGCN & 0.844\small ±0.005 & 0.856 & 0.844 & 0.863 & 0.809 & 0.880\small ±0.009 & 0.793 & 0.912 & 0.947 & 0.849 \\
 ClaHi-GAT & 0.851\small ±0.004 & 0.851 & 0.870 & 0.863 & 0.816 & 0.885\small ±0.010 & 0.798 & 0.952 & 0.917 & 0.854 \\
 GACL & 0.846\small ±0.007 & 0.859 & 0.845 & 0.866 & 0.812 & 0.891\small ±0.004 & 0.802 & 0.929 & 0.945 & 0.872 \\
 DDGCN & 0.835\small ±0.006 & 0.840 & 0.850 & 0.856 & 0.791 & 0.893\small ±0.004 & 0.807 & \textbf{0.931} & 0.946 & 0.871 \\
 RAGCL & 0.862\small ±0.004 & \textbf{0.886} & 0.862 & 0.864 & 0.830 & 0.903\small ±0.003 & 0.834 & 0.921 & \textbf{0.965} & 0.878 \\
 \textbf{P2T3} & \textbf{0.874\small ±0.006} & 0.878 & \textbf{0.894} & \textbf{0.872} & \textbf{0.846} & \textbf{0.911\small ±0.004} & 0.847 & 0.924 & \textbf{0.965} & \textbf{0.892} \\
 \Xhline{1.0pt}
\end{tabular}
}
\caption{Experimental results on Twitter15 and Twitter16 dataset.}
\label{tab:twitter}
\end{table*}

\subsection{Ablation Study}

\subsubsection{Token-wise Embedding}

\begin{table*}[!h]
\centering
\begin{tabular}{lcccc}
 \Xhline{1.0pt}
 \rowcolor{gray!20}
 ~ & \textbf{Weibo} & \textbf{DRWeibo} & \textbf{Twitter15} & \textbf{Twitter16} \\
 \hline
 P2T3 & 0.973 & 0.912 & 0.874 & 0.911 \\
 \hline
 w/o Token-wise Embedding & 0.921(↓0.052) & 0.836(↓0.076) & 0.803(↓0.071) & 0.843(↓0.068) \\
 \quad w/o Chain Identifier & 0.943(↓0.030) & 0.867(↓0.045) &  0.857(↓0.017) & 0.854(↓0.057) \\
 \quad w/o Depth Embedding & 0.952(↓0.021) & 0.893(↓0.019) &  0.867(↓0.007) & 0.876(↓0.035) \\
 \quad w/o Type Embedding & 0.964(↓0.009) & 0.915(↑0.003) &  0.870(↓0.004) & 0.893(↓0.018) \\
 \hdashline
 w/o Pre-training & 0.958(↓0.015) & 0.898(↓0.014) & 0.861(↓0.013) & 0.901(↓0.010) \\
 \Xhline{1.0pt}
\end{tabular}
\caption{Ablation study on token-wise embeddings and model pre-training.}
\label{tab:ab}
\end{table*}

We ran a series of ablation studies to investigate the impact of token-wise embedding and pre-training. We report accuracy in Table~\ref{tab:ab}. The results show that if we directly convert RPT into a sequence without augmenting it for input to Transformer, the model accuracy will be very poor. Among the three token-wise embeddings we used, chain identifier has the greatest impact on model performance, indicating that assisting the model in recognizing the chain structure within the tree is crucial. The type embedding has the smallest impact on the model but still contributes positively to the performance. Additionally, the results also indicate that pre-training on large-scale unlabeled datasets has a significant impact on improving model performance. 

\subsubsection{Model Depth}

\begin{figure}[!h]
  \centering
  \includegraphics[width=0.48\textwidth]{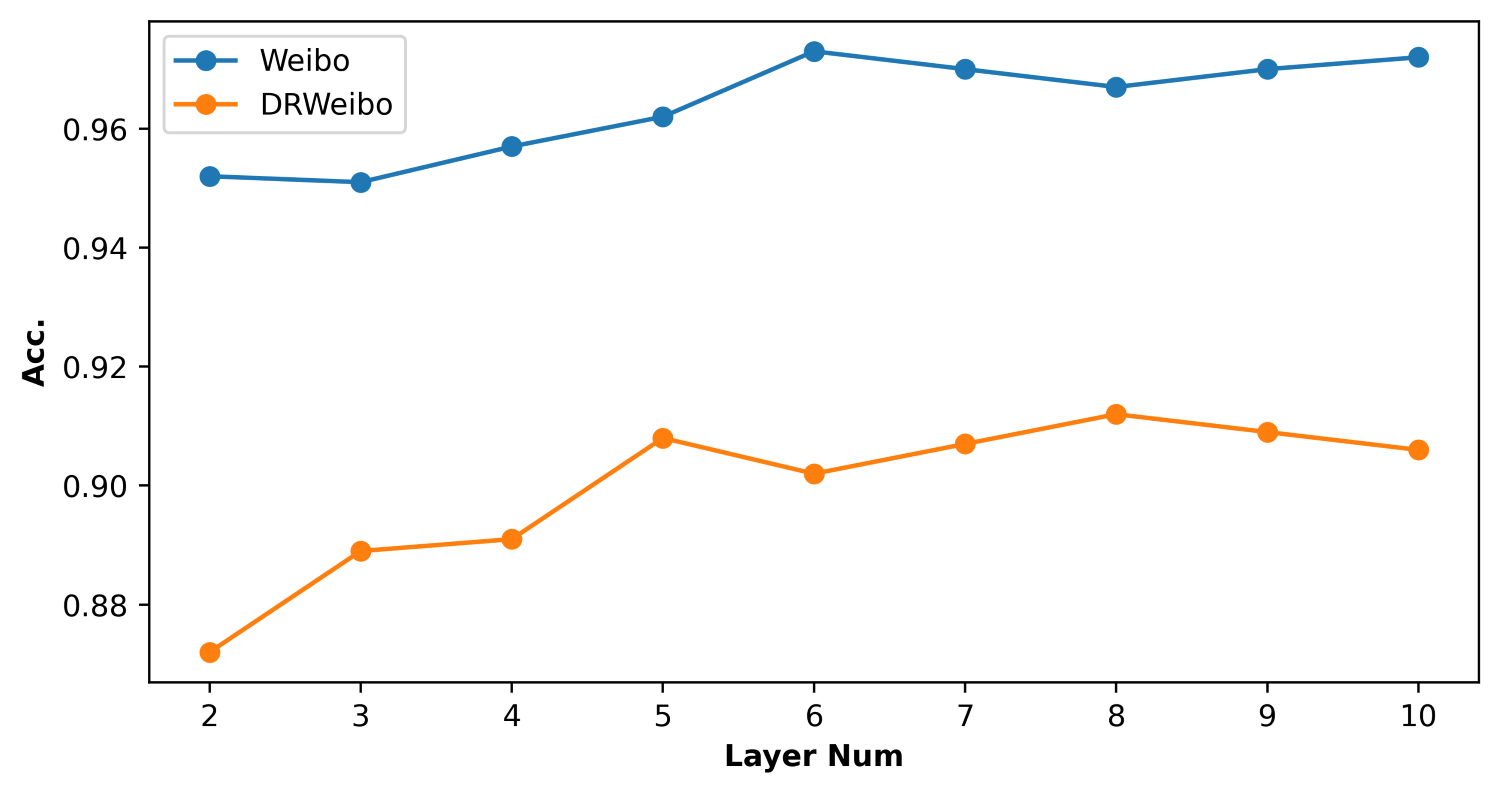}
  \caption{The impact of model layer numbers on P2T3.}
  \label{fig:layernum-p2t3}
\end{figure}

We investigated the impact of model depth on Weibo and DRWeibo in Fig.~\ref{fig:layernum-p2t3}. The results show that as layer number increases, P2T3's performance gradually improves, eventually stabilizing. It shows significant implications for pre-training with a larger model on social media data and P2T3 is not affected by over-smoothing with the increase of model scale. Due to over-fitting and over-smoothing issues associated with GNN-based models, it is challenging to scale up their model size, resulting in limited expressiveness during pre-training process on large-scale datasets. However, as the model size increases, P2T3 achieves superior performance.

\subsection{Pre-training}
\label{sec:abpt}

We further evaluated the impact of pre-training on large-scale data on the performance of different types of models: GIN \cite{gin}, GAT \cite{gat}, GCN \cite{gcn}, BiGCN \cite{bigcn}, RAGCL \cite{ragcl}, and P2T3. Results are shown in Table~\ref{tab:pre}. All experiments utilized the contrastive loss with MI maximization. The results indicate that P2T3 is more prone to benefit from unsupervised pre-training compared to the other models. GNN-based models are constrained by the over-smoothing problem and lack model capacity, making it challenging to effectively utilize large-scale unlabeled data to learn discriminative representations of claims.


\begin{table}[t]
\centering
\resizebox{0.47\textwidth}{!}{
\begin{tabular}{cccc}
 \Xhline{1.0pt}
 \rowcolor{gray!20}
 \textbf{Method} & \textbf{Setting} & \textbf{Weibo} & \textbf{DRWeibo} \\
 \hline
 \multirow{2}{*}{GIN} & w/o Pre-training & 0.940 & 0.847 \\
 ~ & w/ Pre-training & 0.943(↑0.003) & 0.854(↑0.007) \\
 \hline
 \multirow{2}{*}{GAT} & w/o Pre-training & 0.927 & 0.836 \\
 ~ & w/ Pre-training & 0.928(↑0.001) & 0.841(↑0.005) \\
 \hline
 \multirow{2}{*}{GCN} & w/o Pre-training & 0.931 & 0.849 \\
 ~ & w/ Pre-training & 0.939(↑0.008) & 0.851(↑0.002) \\
 \hline
 \multirow{2}{*}{BiGCN} & w/o Pre-training & 0.942 & 0.866 \\
 ~ & w/ Pre-training & 0.940(↓0.002) & 0.871(↑0.005) \\
 \hline
 \multirow{2}{*}{RAGCL} & w/o Pre-training & \textbf{0.960} & 0.896 \\
 ~ & w/ Pre-training & 0.962(↑0.002) & 0.901(↑0.005) \\
 \hline
 \multirow{2}{*}{\textbf{P2T3}} & w/o Pre-training & 0.958 & \textbf{0.898} \\
 ~ & w/ Pre-training & \textbf{0.973(↑0.015)} & \textbf{0.912(↑0.014)} \\
 \Xhline{1.0pt}
\end{tabular}
}
\caption{The impact of pre-training on various models.}
\label{tab:pre}
\end{table}

\subsection{Few-shot Performance}
\label{sec:fs}

\begin{figure}[!h]
  \centering
  \subfigure[Weibo]{\includegraphics[width=0.48\textwidth]{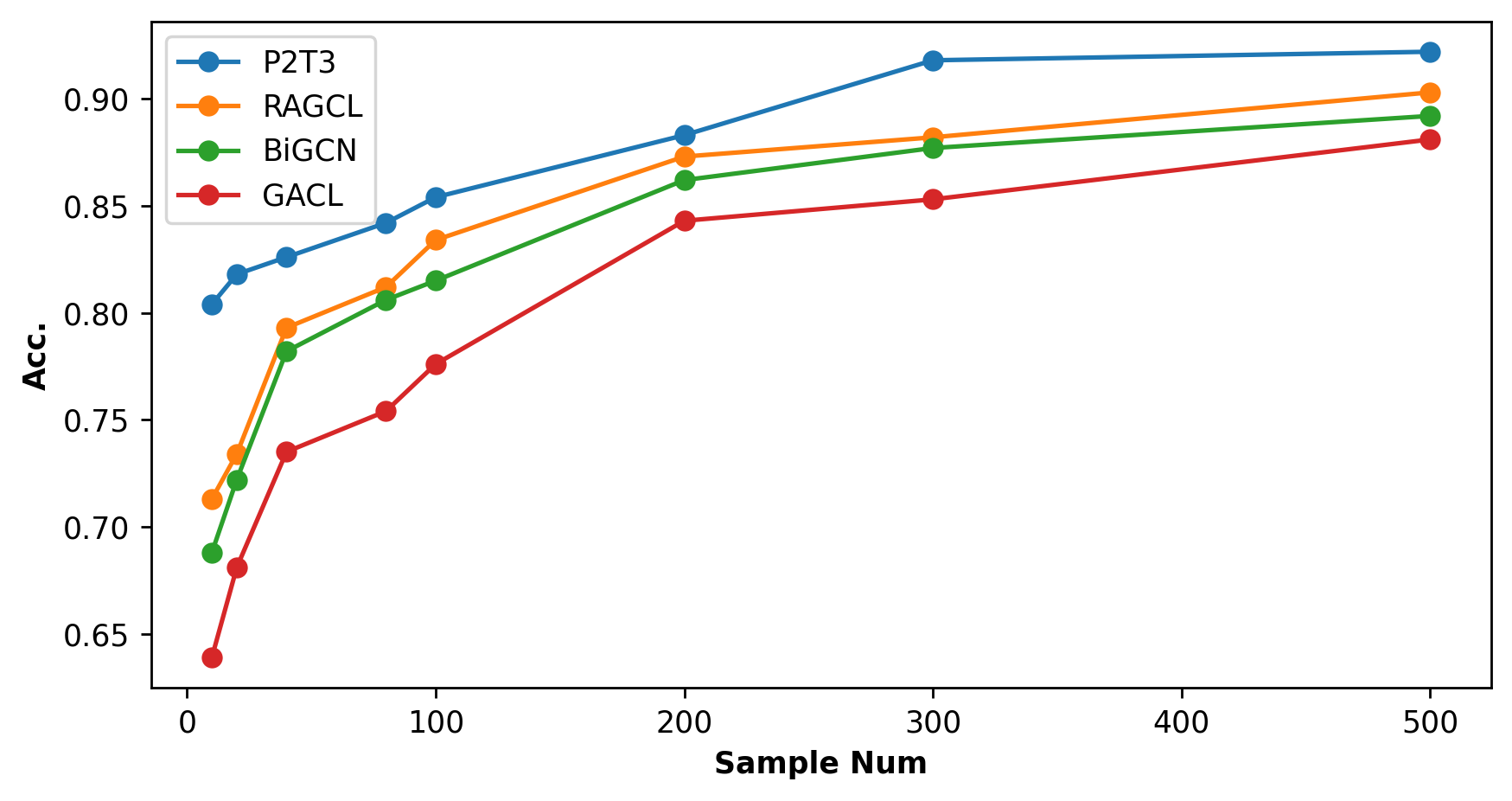}}
  \subfigure[DRWeibo]{\includegraphics[width=0.48\textwidth]{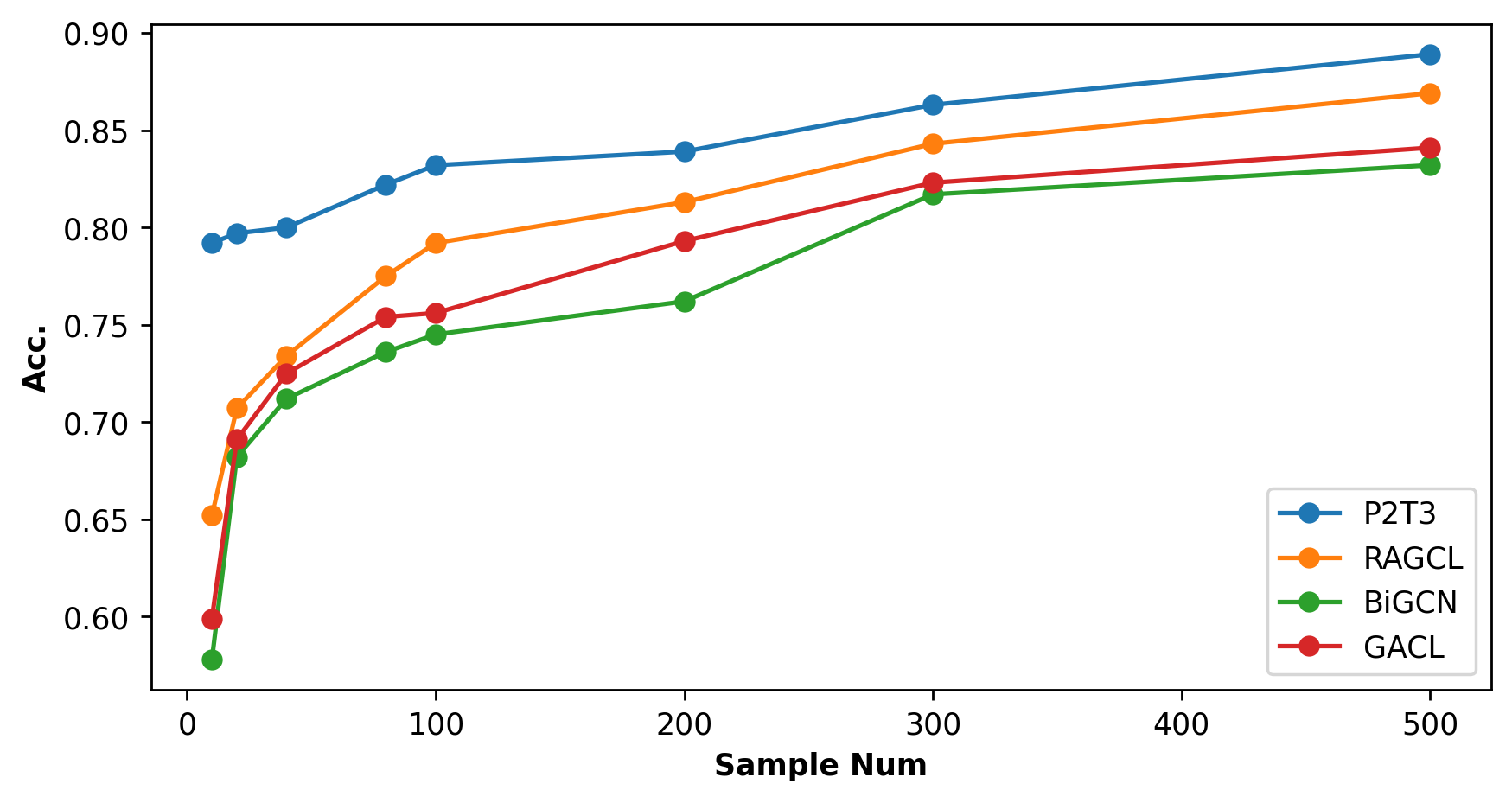}}
  \caption{Results of few-shot experiments.}
  \label{fig:fs}
\end{figure}

We ran few-shot experiments with P2T3, BiGCN, GACL and RAGCL in Fig.~\ref{fig:fs}. We concern the efficacy of these models with minimal labeled samples. This is pertinent given the transient nature of rumors which are often deleted after detection, hampering the acquisition of large-scale labeled datasets. Therefore, if the model can perform effectively in few-shot scenarios, it would reduce the reliance on hard-to-obtain labeled data. We varied the number of labeled samples $k$ between 10 and 500. P2T3 was pre-trained on UWeibo, then fine-tuned using the designated labeled samples. Our results show that P2T3 can leverage unlabeled data to enhance claim representations even in scarce sample conditions, thereby delivering good performance. 

\section{Conclusion}

In recent years, rumor detection methods based on propagation structure learning have become increasingly important in research, with most of these methods employing GNNs as foundational model. However, our investigation indicates that GNNs exacerbate over-smoothing problem when dealing with the unique structure of RPTs. In Addition, the special structure of RPT makes them, compared to other graph structures, more suitable for processing using Transformer architecture. Our P2T3 model leverages the Transformer architecture to handle the distinctive chain-like structure of RPTs while avoiding the over-smoothing issue. This enables P2T3 to harness the rich, unlabeled data resources in social media on a larger network scale.




In future research, we will explore methods to combine P2T3 with large language models and investigate how to utilize P2T3 to construct multi-modal rumor detection models that can handle text attribute graphs and images on social media platforms. Additionally, we aim to extend the application of P2T3 to other social media tasks, such as content recommendation, user behavior analysis, and social network analysis.




\bibliography{custom}

@inproceedings{bigcn,
  title={Rumor detection on social media with bi-directional graph convolutional networks},
  author={Bian, Tian and Xiao, Xi and Xu, Tingyang and Zhao, Peilin and Huang, Wenbing and Rong, Yu and Huang, Junzhou},
  booktitle={Proceedings of the AAAI conference on artificial intelligence},
  volume={34},
  number={01},
  pages={549--556},
  year={2020}
}

@article{ebgcn,
  title={Towards propagation uncertainty: Edge-enhanced bayesian graph convolutional networks for rumor detection},
  author={Wei, Lingwei and Hu, Dou and Zhou, Wei and Yue, Zhaojuan and Hu, Songlin},
  journal={arXiv preprint arXiv:2107.11934},
  year={2021}
}

@inproceedings{gacl,
  title={Rumor Detection on Social Media with Graph Adversarial Contrastive Learning},
  author={Sun, Tiening and Qian, Zhong and Dong, Sujun and Li, Peifeng and Zhu, Qiaoming},
  booktitle={Proceedings of the ACM Web Conference 2022},
  pages={2789--2797},
  year={2022}
}

@article{bert,
  title={Bert: Pre-training of deep bidirectional transformers for language understanding},
  author={Devlin, Jacob and Chang, Ming-Wei and Lee, Kenton and Toutanova, Kristina},
  journal={arXiv preprint arXiv:1810.04805},
  year={2018}
}

@article{gcn,
  title={Semi-supervised classification with graph convolutional networks},
  author={Kipf, Thomas N and Welling, Max},
  journal={arXiv preprint arXiv:1609.02907},
  year={2016}
}

@article{gat,
  title={Graph attention networks},
  author={Veli{\v{c}}kovi{\'c}, Petar and Cucurull, Guillem and Casanova, Arantxa and Romero, Adriana and Lio, Pietro and Bengio, Yoshua},
  journal={arXiv preprint arXiv:1710.10903},
  year={2017}
}

@article{gin,
  title={How powerful are graph neural networks?},
  author={Xu, Keyulu and Hu, Weihua and Leskovec, Jure and Jegelka, Stefanie},
  journal={arXiv preprint arXiv:1810.00826},
  year={2018}
}

@article{graphsage,
  title={Inductive representation learning on large graphs},
  author={Hamilton, Will and Ying, Zhitao and Leskovec, Jure},
  journal={Advances in neural information processing systems},
  volume={30},
  year={2017}
}

@article{dim,
  title={Learning deep representations by mutual information estimation and maximization},
  author={Hjelm, R Devon and Fedorov, Alex and Lavoie-Marchildon, Samuel and Grewal, Karan and Bachman, Phil and Trischler, Adam and Bengio, Yoshua},
  journal={arXiv preprint arXiv:1808.06670},
  year={2018}
}

@article{dgi,
  title={Deep Graph Infomax.},
  author={Velickovic, Petar and Fedus, William and Hamilton, William L and Li{\`o}, Pietro and Bengio, Yoshua and Hjelm, R Devon},
  journal={ICLR (Poster)},
  volume={2},
  number={3},
  pages={4},
  year={2019}
}

@article{infograph,
  title={Infograph: Unsupervised and semi-supervised graph-level representation learning via mutual information maximization},
  author={Sun, Fan-Yun and Hoffmann, Jordan and Verma, Vikas and Tang, Jian},
  journal={arXiv preprint arXiv:1908.01000},
  year={2019}
}

@inproceedings{twitter1516,
  title={Detect rumors in microblog posts using propagation structure via kernel learning},
  author={Ma, Jing and Gao, Wei and Wong, Kam-Fai},
  year={2017},
  organization={Association for Computational Linguistics}
}

@article{weibo,
  title={Detecting rumors from microblogs with recurrent neural networks},
  author={Ma, Jing and Gao, Wei and Mitra, Prasenjit and Kwon, Sejeong and Jansen, Bernard J and Wong, Kam-Fai and Cha, Meeyoung},
  year={2016},
  publisher={AAAI Press}
}

@inproceedings{yucnn,
  title={A Convolutional Approach for Misinformation Identification.},
  author={Yu, Feng and Liu, Qiang and Wu, Shu and Wang, Liang and Tan, Tieniu and others},
  booktitle={IJCAI},
  pages={3901--3907},
  year={2017}
}

@inproceedings{liuandwu,
  title={Early detection of fake news on social media through propagation path classification with recurrent and convolutional networks},
  author={Liu, Yang and Wu, Yi-Fang},
  booktitle={Proceedings of the AAAI conference on artificial intelligence},
  volume={32},
  number={1},
  year={2018}
}

@inproceedings{eann,
  title={Eann: Event adversarial neural networks for multi-modal fake news detection},
  author={Wang, Yaqing and Ma, Fenglong and Jin, Zhiwei and Yuan, Ye and Xun, Guangxu and Jha, Kishlay and Su, Lu and Gao, Jing},
  booktitle={Proceedings of the 24th acm sigkdd international conference on knowledge discovery \& data mining},
  pages={849--857},
  year={2018}
}

@inproceedings{ms1,
  title={Multi-source multi-class fake news detection},
  author={Karimi, Hamid and Roy, Proteek and Saba-Sadiya, Sari and Tang, Jiliang},
  booktitle={Proceedings of the 27th international conference on computational linguistics},
  pages={1546--1557},
  year={2018}
}

@inproceedings{ms2,
  title={A Novel Score-Based Multi-Source Fake News Detection using Gradient Boosting Algorithm},
  author={Birunda, S Selva and Devi, R Kanniga},
  booktitle={2021 International Conference on Artificial Intelligence and Smart Systems (ICAIS)},
  pages={406--414},
  year={2021},
  organization={IEEE}
}

@article{ms3,
  title={Dual co-attention-based multi-feature fusion method for rumor detection},
  author={Bing, Changsong and Wu, Yirong and Dong, Fangmin and Xu, Shouzhi and Liu, Xiaodi and Sun, Shuifa},
  journal={Information},
  volume={13},
  number={1},
  pages={25},
  year={2022},
  publisher={MDPI}
}

@inproceedings{otherrumor1,
  title={Multimodal fusion with recurrent neural networks for rumor detection on microblogs},
  author={Jin, Zhiwei and Cao, Juan and Guo, Han and Zhang, Yongdong and Luo, Jiebo},
  booktitle={Proceedings of the 25th ACM international conference on Multimedia},
  pages={795--816},
  year={2017}
}

@inproceedings{otherrumor2,
  title={Rumor detection by exploiting user credibility information, attention and multi-task learning},
  author={Li, Quanzhi and Zhang, Qiong and Si, Luo},
  booktitle={Proceedings of the 57th annual meeting of the association for computational linguistics},
  pages={1173--1179},
  year={2019}
}

@inproceedings{rvnn,
  title={Rumor detection on twitter with tree-structured recursive neural networks},
  author={Ma, Jing and Gao, Wei and Wong, Kam-Fai},
  year={2018},
  organization={Association for Computational Linguistics}
}

@inproceedings{plan,
  title={Interpretable rumor detection in microblogs by attending to user interactions},
  author={Khoo, Ling Min Serena and Chieu, Hai Leong and Qian, Zhong and Jiang, Jing},
  booktitle={Proceedings of the AAAI conference on artificial intelligence},
  volume={34},
  number={05},
  pages={8783--8790},
  year={2020}
}

@inproceedings{dtc,
  title={Information credibility on twitter},
  author={Castillo, Carlos and Mendoza, Marcelo and Poblete, Barbara},
  booktitle={Proceedings of the 20th international conference on World wide web},
  pages={675--684},
  year={2011}
}

@inproceedings{rfc,
  title={Prominent features of rumor propagation in online social media},
  author={Kwon, Sejeong and Cha, Meeyoung and Jung, Kyomin and Chen, Wei and Wang, Yajun},
  booktitle={2013 IEEE 13th international conference on data mining},
  pages={1103--1108},
  year={2013},
  organization={IEEE}
}

@inproceedings{ddgcn,
  title={Ddgcn: Dual dynamic graph convolutional networks for rumor detection on social media},
  author={Sun, Mengzhu and Zhang, Xi and Zheng, Jiaqi and Ma, Guixiang},
  booktitle={Proceedings of the AAAI conference on artificial intelligence},
  volume={36},
  number={4},
  pages={4611--4619},
  year={2022}
}

@article{transformer,
  title={Attention is all you need},
  author={Vaswani, Ashish and Shazeer, Noam and Parmar, Niki and Uszkoreit, Jakob and Jones, Llion and Gomez, Aidan N and Kaiser, {\L}ukasz and Polosukhin, Illia},
  journal={Advances in neural information processing systems},
  volume={30},
  year={2017}
}

@article{tfg,
  title={Transformer for graphs: An overview from architecture perspective},
  author={Min, Erxue and Chen, Runfa and Bian, Yatao and Xu, Tingyang and Zhao, Kangfei and Huang, Wenbing and Zhao, Peilin and Huang, Junzhou and Ananiadou, Sophia and Rong, Yu},
  journal={arXiv preprint arXiv:2202.08455},
  year={2022}
}

@article{rsa1,
  title={A generalization of transformer networks to graphs},
  author={Dwivedi, Vijay Prakash and Bresson, Xavier},
  journal={arXiv preprint arXiv:2012.09699},
  year={2020}
}

@inproceedings{rsa2,
  title={Universal Graph Transformer Self-Attention Networks},
  author={Nguyen, Tu Dinh and Phung, Dinh and others},
  booktitle={International World Wide Web Conference 2022},
  pages={193--196},
  year={2022},
  organization={Association for Computing Machinery (ACM)}
}

@inproceedings{amp2,
  title={Mesh graphormer},
  author={Lin, Kevin and Wang, Lijuan and Liu, Zicheng},
  booktitle={Proceedings of the IEEE/CVF international conference on computer vision},
  pages={12939--12948},
  year={2021}
}

@inproceedings{amp3,
  title={Specformer: Spectral Graph Neural Networks Meet Transformers},
  author={Bo, Deyu and Shi, Chuan and Wang, Lele and Liao, Renjie},
  booktitle={The Eleventh International Conference on Learning Representations},
  year={2023}
}

@inproceedings{oversm1,
  title={Deeper insights into graph convolutional networks for semi-supervised learning},
  author={Li, Qimai and Han, Zhichao and Wu, Xiao-Ming},
  booktitle={Proceedings of the AAAI conference on artificial intelligence},
  volume={32},
  number={1},
  year={2018}
}

@article{oversm2,
  title={A note on over-smoothing for graph neural networks},
  author={Cai, Chen and Wang, Yusu},
  journal={arXiv preprint arXiv:2006.13318},
  year={2020}
}

@article{oversm3,
  title={Graph neural networks exponentially lose expressive power for node classification},
  author={Oono, Kenta and Suzuki, Taiji},
  journal={arXiv preprint arXiv:1905.10947},
  year={2019}
}

@article{oversm4,
  title={Revisiting graph neural networks: All we have is low-pass filters},
  author={Nt, Hoang and Maehara, Takanori},
  journal={arXiv preprint arXiv:1905.09550},
  year={2019}
}

@article{oversm5,
  title={Not too little, not too much: a theoretical analysis of graph (over) smoothing},
  author={Keriven, Nicolas},
  journal={Advances in Neural Information Processing Systems},
  volume={35},
  pages={2268--2281},
  year={2022}
}

@article{oversm6,
  title={Beyond homophily in graph neural networks: Current limitations and effective designs},
  author={Zhu, Jiong and Yan, Yujun and Zhao, Lingxiao and Heimann, Mark and Akoglu, Leman and Koutra, Danai},
  journal={Advances in neural information processing systems},
  volume={33},
  pages={7793--7804},
  year={2020}
}

@article{hm1,
  title={Do transformers really perform badly for graph representation?},
  author={Ying, Chengxuan and Cai, Tianle and Luo, Shengjie and Zheng, Shuxin and Ke, Guolin and He, Di and Shen, Yanming and Liu, Tie-Yan},
  journal={Advances in Neural Information Processing Systems},
  volume={34},
  pages={28877--28888},
  year={2021}
}

@inproceedings{hm3,
  title={Grpe: Relative positional encoding for graph transformer},
  author={Park, Wonpyo and Chang, Woong-Gi and Lee, Donggeon and Kim, Juntae and others},
  booktitle={ICLR2022 Machine Learning for Drug Discovery},
  year={2022}
}

@article{hm5,
  title={Sign and basis invariant networks for spectral graph representation learning},
  author={Lim, Derek and Robinson, Joshua and Zhao, Lingxiao and Smidt, Tess and Sra, Suvrit and Maron, Haggai and Jegelka, Stefanie},
  journal={arXiv preprint arXiv:2202.13013},
  year={2022}
}

@article{hm6,
  title={ABT-MPNN: an atom-bond transformer-based message-passing neural network for molecular property prediction},
  author={Liu, Chengyou and Sun, Yan and Davis, Rebecca and Cardona, Silvia T and Hu, Pingzhao},
  journal={Journal of Cheminformatics},
  volume={15},
  number={1},
  pages={29},
  year={2023},
  publisher={Springer}
}

@inproceedings{hm7,
  title={Hierarchical transformer for scalable graph learning},
  author={Zhu, Wenhao and Wen, Tianyu and Song, Guojie and Ma, Xiaojun and Wang, Liang},
  booktitle={Proceedings of the Thirty-Second International Joint Conference on Artificial Intelligence},
  pages={4702--4710},
  year={2023}
}

@article{tokengt,
  title={Pure transformers are powerful graph learners},
  author={Kim, Jinwoo and Nguyen, Dat and Min, Seonwoo and Cho, Sungjun and Lee, Moontae and Lee, Honglak and Hong, Seunghoon},
  journal={Advances in Neural Information Processing Systems},
  volume={35},
  pages={14582--14595},
  year={2022}
}

@inproceedings{long,
  title={Representation learning on graphs with jumping knowledge networks},
  author={Xu, Keyulu and Li, Chengtao and Tian, Yonglong and Sonobe, Tomohiro and Kawarabayashi, Ken-ichi and Jegelka, Stefanie},
  booktitle={International conference on machine learning},
  pages={5453--5462},
  year={2018},
  organization={PMLR}
}

@article{rope,
  title={Roformer: Enhanced transformer with rotary position embedding},
  author={Su, Jianlin and Lu, Yu and Pan, Shengfeng and Murtadha, Ahmed and Wen, Bo and Liu, Yunfeng},
  journal={arXiv preprint arXiv:2104.09864},
  year={2021}
}

@article{adam,
  title={Adam: A method for stochastic optimization},
  author={Kingma, Diederik P and Ba, Jimmy},
  journal={arXiv preprint arXiv:1412.6980},
  year={2014}
}

@article{bottleneck,
  title={On the bottleneck of graph neural networks and its practical implications},
  author={Alon, Uri and Yahav, Eran},
  journal={arXiv preprint arXiv:2006.05205},
  year={2020}
}

@article{oversqu1,
  title={Understanding over-squashing and bottlenecks on graphs via curvature},
  author={Topping, Jake and Di Giovanni, Francesco and Chamberlain, Benjamin Paul and Dong, Xiaowen and Bronstein, Michael M},
  journal={arXiv preprint arXiv:2111.14522},
  year={2021}
}

@inproceedings{oversqu2,
  title={Expander graph propagation},
  author={Deac, Andreea and Lackenby, Marc and Veli{\v{c}}kovi{\'c}, Petar},
  booktitle={Learning on Graphs Conference},
  pages={38--1},
  year={2022},
  organization={PMLR}
}

@inproceedings{ragcl,
  title={Propagation Tree Is Not Deep: Adaptive Graph Contrastive Learning Approach for Rumor Detection},
  author={Cui, Chaoqun and Jia, Caiyan},
  booktitle={Proceedings of the AAAI Conference on Artificial Intelligence},
  volume={38},
  number={1},
  pages={73--81},
  year={2024}
}

@inproceedings{treetrans,
  title={Debunking rumors on twitter with tree transformer},
  author={Ma, Jing and Gao, Wei},
  year={2020},
  organization={ACL}
}

@article{clahi,
  title={Rumor detection on twitter with claim-guided hierarchical graph attention networks},
  author={Lin, Hongzhan and Ma, Jing and Cheng, Mingfei and Yang, Zhiwei and Chen, Liangliang and Chen, Guang},
  journal={arXiv preprint arXiv:2110.04522},
  year={2021}
}

@article{25ts1,
  title={Rumor Detection by Multi-task Suffix Learning based on Time-series Dual Sentiments},
  author={Liu, Zhiwei and Yang, Kailai and Hovy, Eduard and Ananiadou, Sophia},
  journal={arXiv preprint arXiv:2502.14383},
  year={2025}
}

@inproceedings{25pl1,
  title={Rumor Detection on Social Media with Temporal Propagation Structure Optimization},
  author={Peng, Xingyu and Wu, Junran and Liu, Ruomei and Xu, Ke},
  booktitle={Proceedings of the 31st International Conference on Computational Linguistics},
  pages={3865--3878},
  year={2025}
}

@article{25mm1,
  title={A unified framework for multi-modal rumor detection via multi-level dynamic interaction with evolving stances},
  author={Sun, Tiening and Liu, Chengwei and Chen, Lizhi and Qian, Zhong and Li, Peifeng and Zhu, Qiaoming},
  journal={Information Processing \& Management},
  volume={62},
  number={3},
  pages={104066},
  year={2025},
  publisher={Elsevier}
}

@inproceedings{adgscl,
  title={Towards Real-World Rumor Detection: Anomaly Detection Framework with Graph Supervised Contrastive Learning},
  author={Cui, Chaoqun and Jia, Caiyan},
  booktitle={Proceedings of the 31st International Conference on Computational Linguistics},
  pages={7141--7155},
  year={2025}
}

@inproceedings{pep,
  title={Enhancing Rumor Detection Methods with Propagation Structure Infused Language Model},
  author={Cui, Chaoqun and Li, Siyuan and Ma, Kunkun and Jia, Caiyan},
  booktitle={Proceedings of the 31st International Conference on Computational Linguistics},
  pages={7165--7179},
  year={2025}
}

@article{urumor,
  title={Graph representation learning with massive unlabeled data for rumor detection},
  author={Cui, Chaoqun and Jia, Caiyan},
  journal={arXiv preprint arXiv:2508.04252},
  year={2025}
}

@article{25os1,
  title={The Oversmoothing Fallacy: A Misguided Narrative in GNN Research},
  author={Park, MoonJeong and Choi, Sunghyun and Heo, Jaeseung and Park, Eunhyeok and Kim, Dongwoo},
  journal={arXiv preprint arXiv:2506.04653},
  year={2025}
}

@article{25os2,
  title={On vanishing gradients, over-smoothing, and over-squashing in gnns: Bridging recurrent and graph learning},
  author={Arroyo, {\'A}lvaro and Gravina, Alessio and Gutteridge, Benjamin and Barbero, Federico and Gallicchio, Claudio and Dong, Xiaowen and Bronstein, Michael and Vandergheynst, Pierre},
  journal={arXiv preprint arXiv:2502.10818},
  year={2025}
}

@article{25os3,
  title={Rethinking Over-Smoothing in Graph Neural Networks: A Perspective from Anderson Localization},
  author={Ouyang, Kaichen},
  journal={arXiv preprint arXiv:2507.05263},
  year={2025}
}

@article{25gt1,
  title={Transformers are graph neural networks},
  author={Joshi, Chaitanya K},
  journal={arXiv preprint arXiv:2506.22084},
  year={2025}
}

@article{25gt2,
  title={Relational Graph Transformer},
  author={Dwivedi, Vijay Prakash and Jaladi, Sri and Shen, Yangyi and L{\'o}pez, Federico and Kanatsoulis, Charilaos I and Puri, Rishi and Fey, Matthias and Leskovec, Jure},
  journal={arXiv preprint arXiv:2505.10960},
  year={2025}
}

@article{25gt3,
  title={T-graphormer: Using transformers for spatiotemporal forecasting},
  author={Bai, Hao Yuan and Liu, Xue},
  journal={arXiv preprint arXiv:2501.13274},
  year={2025}
}

@inproceedings{25gt4,
  title={GRAPH GENERATIVE PRE-TRAINED TRANSFORMER},
  author={Chen, Xiaohui and Wang, Yinkai and He, Jiaxing and Du, Yuanqi and Hassoun, Soha and Xu, Xiaolin and Liu, Liping},
  booktitle={ICLR 2025 Workshop on Deep Generative Model in Machine Learning: Theory, Principle and Efficacy}
}

@article{vga,
  title={VGA: vision and graph fused attention network for rumor detection},
  author={Bai, Lin and Jia, Caiyan and Song, Ziying and Cui, Chaoqun},
  journal={ACM Transactions on Knowledge Discovery from Data},
  volume={19},
  number={4},
  pages={1--21},
  year={2025},
  publisher={ACM New York, NY}
}

@article{debiased,
  title={A debiased self-training framework with graph self-supervised pre-training aided for semi-supervised rumor detection},
  author={Qiao, Yuhan and Cui, Chaoqun and Wang, Yiying and Jia, Caiyan},
  journal={Neurocomputing},
  volume={604},
  pages={128314},
  year={2024},
  publisher={Elsevier}
}

\appendix

\section{Conversation Chain Examples}
\label{sec:cc}

\begin{figure}[!h]
  \centering
  \subfigure[Rumor]{\includegraphics[width=0.48\linewidth]{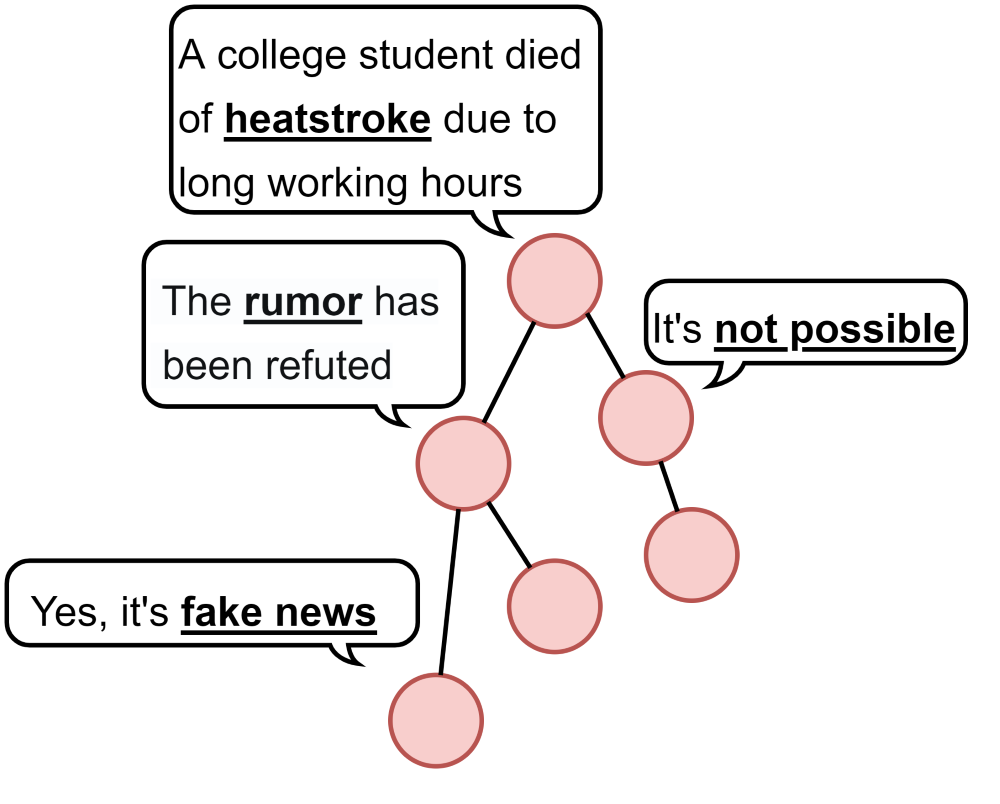}}
  \subfigure[Non-rumor]{\includegraphics[width=0.48\linewidth]{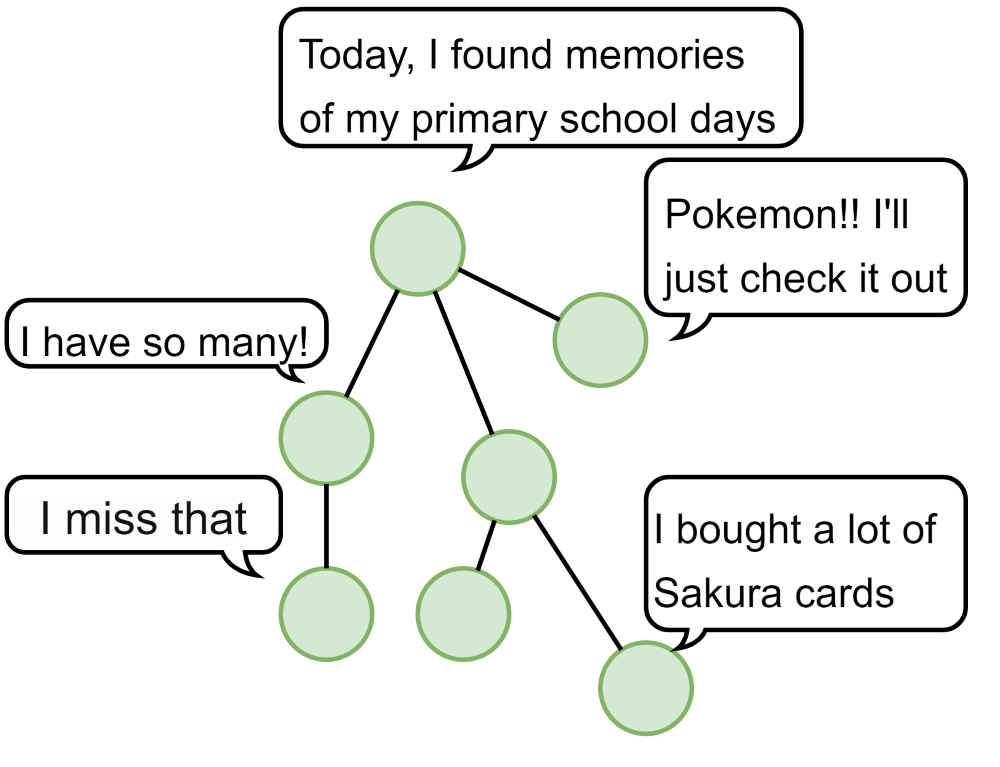}}
  \caption{Examples of RPTs. Comments under rumor thread typically express more heated stances and sentiments.}
  \label{fig:pt}
\end{figure}

Two examples of rumor propagation trees are shown in Figure~\ref{fig:pt}. In Tables~\ref{tab:cc}, we present several conversation chains with their source posts. It's evident that these chains, often chronologically arranged, exhibit clear semantic relations and occasionally intense emotional expressions. These are notable features for identifying rumors.


We also present a case study in Tables~\ref{tab:cc}, where we rank the replies in the conversation chains according to the self-attention scores (averaged across multiple heads) from the last layer of the encoder in P2T3. This helps us investigate which types of replies the model focuses on. The ranking results indicate that replies expressing strong stances and sentiments—rather than those that are simply longer or shorter—receive higher self-attention scores across the three conversation chain examples. Notably, these replies sometimes appear at deeper levels of the conversation chain, which can pose challenges for GNNs during learning. However, the Transformer architecture effectively attends to these deeper nodes. Therefore, P2T3 can leverage the Transformer architecture to learn discriminative patterns for rumor detection from these long conversation chains.


\begin{table*}[!h]
\centering
\begin{tabularx}{\textwidth}{cXc}
 \Xhline{1.0pt}
 \rowcolor{gray!20}
 \textbf{Depth} & \textbf{Post} & \textbf{Rank} \\
 \hline
 source & JUST IN: Donald Trump to @MSNBC: `There has to be some form of punishment' for women who have an abortion. & - \\
 \hdashline
 1 & \textcolor{red}{UPDATE}: Donald Trump advocates abortion ban and \textcolor{red}{`some form of punishment' for women} who have an abortion. & 3\\
 \hdashline
 2 & \textcolor{red}{NEW}: Trump \textcolor{red}{has released this statement} following his abortion comments to @MSNBC. & 4\\
 \hdashline
 3 & In new statement, Trump says doctors, or anyone performing abortion, would be held legally responsible under a ban -- \textcolor{red}{`not the woman.'} & 2\\
 \hdashline
 4 & ok \textcolor{red}{you lost me here!} I've just rtweeted he said women shld get some punushment 4 that & 1\\
 \hline
 source & Attorney: New audio reveals pause in gunfire when Michael Brown was shot. & - \\
 \hdashline
 1 & That's because Brown DIDN'T STOP and he KEPT CHARGING. Idiots are going to be like `OMG HE STOOD OVER HIM AND EXECUTED HIM!' Ugh. & 6\\
 \hdashline
 2 & I understand your intolerance for idiots, but what gives you the impression that your version is \textcolor{red}{absolutely correct?} & 3\\
 \hdashline
 3 & The independent witness being the guy who DOESN'T have a history of falsifying police reports and DOESN'T know Brown. & 9\\
 \hdashline
 4 & So you are discounting the other five witnesses? \textcolor{red}{Falsifying reports?} Who is this independent one? I have not heard of them & 7\\
 \hdashline
 5 & ... as opposed to people who were told, `Stand in front of my camera and tell me what you saw.' & 8\\
 \hdashline
 6 & Do you throw away every witness that ever goes on TV? That seems like a \textcolor{red}{silly measure for integrity}. & 4\\
 \hdashline
 7 & Do you believe everything you see on TV? & 5\\
 \hdashline
 8 & There is another, \textcolor{red}{very credible witness}, but I am having trouble on my phone. The audio is interesting but \textcolor{red}{is it credible?} & 1 \\
 \hdashline
 9 & It's \textcolor{red}{credible}, but it's currently as \textcolor{red}{unverified} as any of those so-called eyewitness reports. FBI is analyzing. & 2\\
 \hline
 source & NBC: arrest records show \#Ferguson cops jailed twice as many people last night as they originally claimed. & -\\
 \hdashline
 1 & meanwhile, \#Ferguson officer suffered fractured eye socket when attacked by Michael Brown. & 6\\
 \hdashline
 2 & \textcolor{red}{prove} thats the cops xray? & 5\\
 \hdashline
 3 & \textcolor{red}{it clearly says} that it's a file image. Arrow shows where that type of fracture occurs. & 2\\
 \hdashline
 4 & so in other words..... \textcolor{red}{NOT the cops}. \textcolor{red}{Zero proof} he was injured at all. If he is so innocent why hide. & 3\\
 \hdashline
 5 & fact that officer was injured \textcolor{red}{was mentioned in the news}. & 4\\
 \hdashline
 6 & you don't get it. \textcolor{red}{They LIE. The media lies. The cops lie. Lies lies lies.} Give me proof. A video? Proof & 1 \\
 \Xhline{1.0pt}
\end{tabularx}
\caption{Conversation chain examples. Segments in tweets that express strong stances or sentiments are highlighted in red.}
\label{tab:cc}
\end{table*}

\section{Related Work}

In this section, we will review the related works on rumor detection and graph Transformers.



\subsection{Social Media Rumor Detection}

Among the existing studies, early rumor detection methods mainly take advantage of traditional classification methods by using hand-crafted features \cite{dtc,rfc}. Deep learning has greatly promoted the development of rumor detection approaches. These approaches can be broadly categorized into four classes, including time-series based methods \cite{yucnn,liuandwu,25ts1} which model text content or user profiles as time series, propagation structure learning methods \cite{ebgcn,ragcl,debiased,25pl1,adgscl,urumor} which consider the propagation structures of rumors, multi-source integation methods \cite{ms1,ms2,ms3,pep} which combine multiple resources of rumors including post content, user profiles, heterogeneous relations between posts and users, multi-modal fusion methods \cite{eann,otherrumor2,25mm1,vga} which incorporate both post and related images to effectively debunk rumors. 

The significance of propagation structure has been increasingly recognized in the research. Many SOTA models bank on learning representations of RPTs using GNNs. 
RvNN \cite{rvnn} designed a bottom-up and top-down tree-structured recursive neural network to extract information from RPTs. 
PLAN \cite{plan} constructed a Transformer model that was aware of the RPT structure. 
HD-TRANS   utilized subtree attention based on Transformer to aggregate neighborhood information.
BiGCN \cite{bigcn} applied a bidirectional GCN alongside a root node feature enhancement technique. 
ClaHi-GAT \cite{clahi} utilized an undirected graph that integrates sibling relations in conjunction with GAT to model user interactions.
GACL \cite{gacl} incorporated contrastive loss with adversarial training to learn representations robust to noise. 
DDGCN \cite{ddgcn} modeled multiple types of information in one unified framework.
Recently, RAGCL \cite{ragcl} utilized adaptive contrastive learning to cope with the imbalance of RPTs.
These studies demonstrate the effectiveness of propagation structure learning.

\subsection{Over-smoothing and Graph Transformers}


The number of layers in a neural network (referred to as depth) is often considered to be crucial for its performance on real-world tasks. For example, convolutional neural networks (CNNs) used in computer vision, often use tens or even hundreds of layers. In contrast, most GNNs encountered in applications are relatively shallow and often have just few layers. This is related to several issues impairing the performance of deep GNNs in realistic graph learning settings: graph bottlenecks \cite{bottleneck}, over-squashing \cite{oversqu1,oversqu2,25os2}, and over-smoothing \cite{oversm1,oversm3,oversm4,25os1}. In this study we focus on the over-smoothing phenomenon, which loosely refers to the exponential convergence of all node features towards the same constant value as the number of layers in the GNN increases. While it has been shown that small amounts of smoothing are desirable for regression and classification tasks \cite{oversm5}, excessive smoothing (or over-smoothing) results in convergence to a non-informative limit. Besides being a key limitation in the development of deep multi-layer GNNs, over-smoothing can also severely impact the ability of GNNs to handle heterophilic graphs \cite{oversm6,25os3}, in which node labels tend to differ from the labels of the neighbors and thus long-term interactions have to be learned.

Applying Transformers to graph is challenging mainly due to (1) the presence of edge connectivity and (2) the absence of a canonical node ordering, which makes the adoption of simple positional encodings unfeasible \cite{tfg}. To address the issue of edge connectivity, early methods constrained self-attention to local neighborhoods, effectively reducing it to message passing \cite{rsa1,rsa2}. Alternatively, global self-attention has been employed alongside auxiliary message passing modules to account for edge connectivity \cite{amp2,amp3,25gt1,25gt2}. However, message-passing methods suffer from limited expressive power \cite{gin} and over-smoothing issues \cite{oversm3,oversm2,25os2}, leading recent works to discard them in favor of global self-attention on nodes. To process edges, heuristic modifications are often introduced \cite{hm1,hm3,hm5,hm6,hm7,25gt3,25gt4}. These adaptations aim to overcome the challenges posed by graph structures in order to enhance the Transformers performance in graph-related tasks. 


\section{Experiment Details}
\label{sec:expdetail}

In this section, we mainly introduce some specific settings in our experiments, including the way of data preprocessing and the hyperparameter configuration when training the model.

\subsection{Unlabeled Dataset Construction}
\label{sec:databuild}

For the UWeibo dataset, we employed web crawler techniques to randomly collect trending posts and their complete propagation structures from the homepage of popular Weibo posts\footnote{\url{https://weibo.com/hot/weibo/102803}}. To ensure the dataset's integrity and independence from platform recommendation algorithms, we utilized multiple newly created accounts to extract data. This approach aimed to mitigate potential biases that might arise from the platform's algorithms and to reflect the genuine domain distribution of social media content. Regarding UTwitter dataset, we initially utilized multiple newly created accounts to randomly follow high-follower count influencers. Subsequently, we conducted random crawling of posts and their propagation structures from the Twitter homepage\footnote{\url{https://twitter.com/home}}. Due to the fact that UTwitter dataset is exclusively sourced from users with a substantial number of followers, the authenticity of the posts is more likely to be ensured. The code for the web scraping program can be found at \url{https://anonymous.4open.science/r/WeiboCrawl-64D2} and \url{https://anonymous.4open.science/r/TwitterCrawl-BE29}.

Due to the stringent regulation imposed by platforms on the dissemination of rumors, acquiring a sufficiently large-scale labeled dataset for rumor detection proves to be exceptionally challenging. Conversely, obtaining extensive amounts of unlabeled data is relatively simpler, especially with the availability of platform data APIs offered by certain mainstream social media platforms (e.g., Twitter API). Consequently, we posit that future research should place greater emphasis on semi-supervised rumor detection methods.

\subsection{Data Preprocessing}

For the texts in all datasets, we first standardize the different fonts present in the texts, then identify user mentions and web/url links as special tokens, \texttt{<@user>} and \texttt{<url>}. Next, we use the TweetTokenizer from the NLTK toolkit and jieba word segmentation engine to tokenize the raw texts in English and Chinese datasets, respectively. Additionally, we use the \texttt{emoji} package\footnote{\url{https://pypi.org/project/emoji}} to translate the emojis in the texts into text string tokens.

\subsection{Hyperparameter Configuration}

All models are implemented by PyTorch and baseline methods are re-implemented. GACL uses BERT \cite{bert} to extract initial feature vector of each post in RPTs. In addition to GACL, other models use 200-dimensional word2vec word embeddings as initial feature vectors. In the main experiments of P2T3, we set batch size to 32, learning rate to 5e-5. The Transformer encoder consists of 3 layers. We optimize the loss function using the Adam optimizer \cite{adam}. The entire training process is conducted on a single Nvidia GeForce RTX 3090 GPU.

BiGCN and GACL utilize early stopping to observe the performance. However, due to oscillations in the early stages of model training, the observed model performance is unstable. In order to compare the performance of different models more fairly, we conduct experiments on P2T3 and multiple baseline methods with the same data, while all models are trained for 100 epochs until convergence. We consider the average results of the final 10 epochs out of these 100 as the stable outcome that the models can achieve.

\end{document}